\documentclass[10pt,twocolumn,letterpaper]{article}

\usepackage{wacv}              

\usepackage{graphicx}
\usepackage{amsmath}
\usepackage{amssymb}
\usepackage{booktabs}
\usepackage{times}
\usepackage{epsfig}
\usepackage{graphicx}
\usepackage{amsmath}
\usepackage{amssymb}
\usepackage{multirow}
\usepackage{tabularx}
\usepackage{colortbl}
\usepackage{xcolor}
\usepackage{amsmath}
\usepackage{algorithm}
\usepackage{algpseudocode}
\usepackage{subcaption}
\usepackage{caption}
\usepackage{booktabs}
\usepackage{multirow}
\usepackage{hhline}
\usepackage[accsupp]{axessibility}

\usepackage[pagebackref,breaklinks,colorlinks]{hyperref}

\usepackage[capitalize]{cleveref}
\crefname{section}{Sec.}{Secs.}
\Crefname{section}{Section}{Sections}
\Crefname{table}{Table}{Tables}
\crefname{table}{Tab.}{Tabs.}

\def\wacvPaperID{915} 
\def\confName{WACV}
\def\confYear{2025}

\begin{document}

\title{ColFigPhotoAttnNet: Reliable Finger Photo Presentation Attack Detection Leveraging Window-Attention on Color Spaces}

\author{
Anudeep Vurity\textsuperscript{1}, Emanuela Marasco\textsuperscript{1}, Raghavendra Ramachandra\textsuperscript{2}, Jongwoo Park\textsuperscript{3} \\
\\
\textsuperscript{1}Center for Secure Information Systems, George Mason University, U.S.A.\\
\textsuperscript{2}Norwegian University of Science and Technology (NTNU), Gjøvik, Norway\\
\textsuperscript{3}Stony Brook University, Stony Brook, NY, U.S.A\\
{\tt\small \{avurity, emarasco\}@gmu.edu, raghavendra.ramachandra@ntnu.no, jongwopark@cs.stonybrook.edu}
}

\maketitle

\begin{abstract}
Finger photo Presentation Attack Detection (PAD) can significantly strengthen smartphone device security. However, these algorithms are trained to detect certain types of attacks. Furthermore, they are designed to operate on images acquired by specific capture devices, leading to poor generalization and a lack of robustness in handling the evolving nature of mobile hardware.
The proposed investigation is the first to systematically analyze the performance degradation of existing deep learning PAD systems, convolutional and transformers, in cross-capture device settings. In this paper, we introduce the \textit{ColFigPhotoAttnNet}\footnote{The implementation of ColFigPhotoAttnNet can be accessed at \url{https://github.com/avurity/ColFigPhotoAttnNet}} architecture designed based on window attention on color channels, followed by the nested residual network as the predictor to achieve a reliable PAD. 
Extensive experiments using various capture devices, including iPhone13 Pro, GooglePixel 3, Nokia C5, and OnePlusOne, were carried out to evaluate the performance of proposed and existing methods on three publicly available databases. The findings underscore the effectiveness of our approach. 
\end{abstract}

\section{Introduction}
The demand for robust security and seamless user identity verification is allowing growth in mobile biometric adoption.
Smartphone cameras enable contactless biometric fingerprint capture for reliable identity verification in various practical applications, including mobile voting, BFSI (Banking, Financial Services, and Insurance), and healthcare sectors \cite{wild2019comparative,gentles2012application,aratekOverviewMobile}.
Government initiatives promote the integration of these technologies into public services and enterprises. Through the Mobile Biometric Application (MBA) program, FBI agents and federal task force officers (TFOs) use mobile devices (e.g., cell phones and tablets) to confirm an individual's identity in situations and locations where mobile biometric identification is necessary, such as mass arrests and natural disasters.

\begin{figure}[]
\centering
{\includegraphics[width=8.4cm,height=6.2cm]{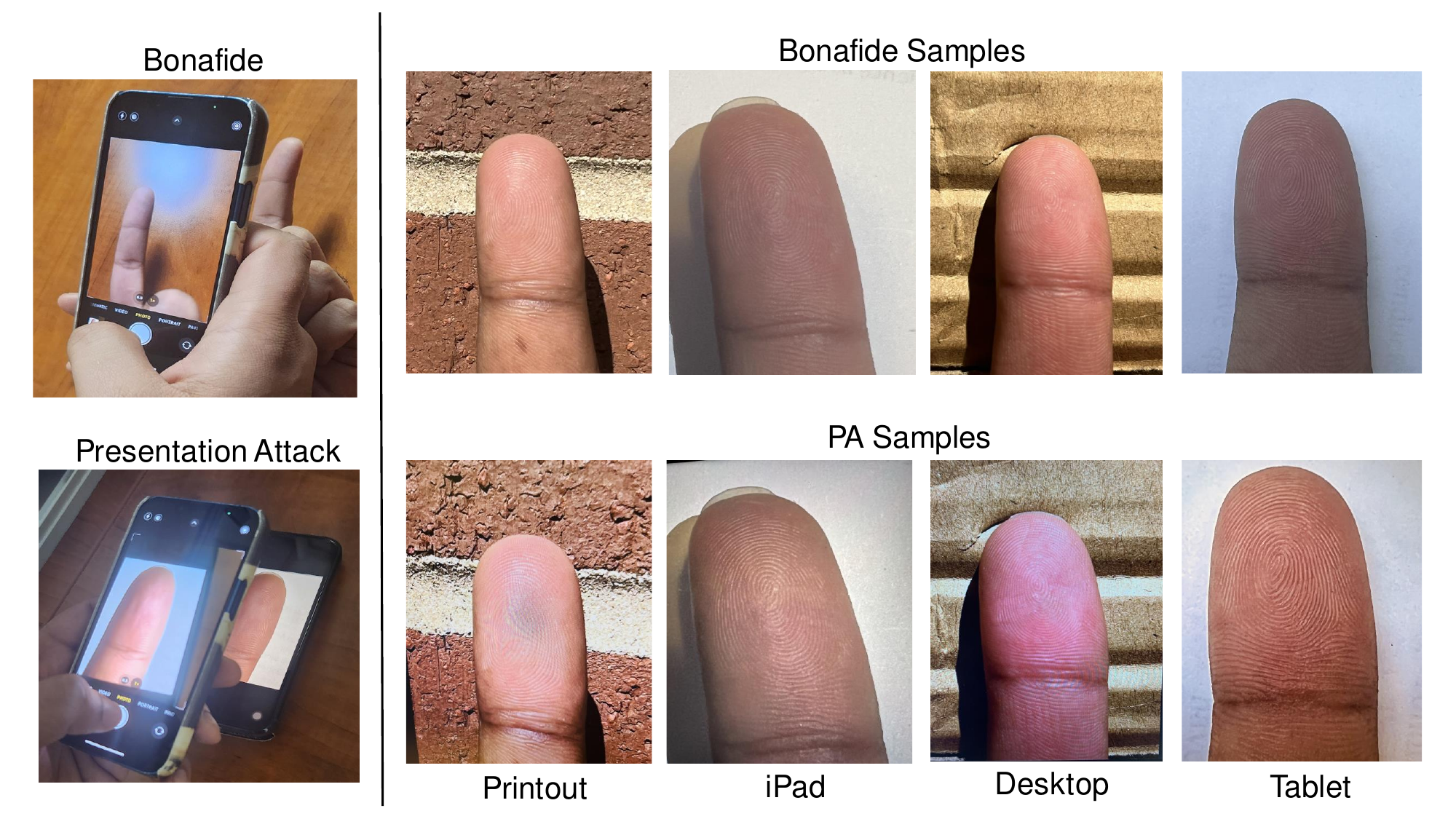}}
\caption{On the left, it shows the bona fide and attack capture mechanisms. On the right, there are examples of each scenario.}
\centering
\label{intro}
\end{figure}

Although finger photo recognition can effectively secure these devices, the technology is vulnerable to presentation attacks (PAs). Examples of bona fide and PAs are illustrated in Figure \ref{intro}. 
Although PAD algorithms can fortify the system against these threats, they are not designed to handle devices' fast innovation and continuous evolution \cite{sauter2013evolution,fabbrizzi2022survey}.

Modern smartphones have changed the photography function significantly, with upgraded cameras featuring high-definition, night mode, and anti-shake characteristics. These changes contribute to PAD performance degradation. In critical applications impacting human beings, creating trustworthy decision-making systems is essential \cite{marasco2019biases,shahbazi2023representation,marasco2022fingerpin}. Capture bias is related to how the images are acquired both in terms of the used device and the collector's preferences regarding point of view, lighting conditions, etc. \cite{tommasi2017deeper,torralba2011unbiased,khosla2012undoing}. If not addressed, this concern can increase mistrust of the technology \cite{jain2021biometrics}.

The development of attention mechanisms brought many benefits to several computer vision tasks and was inspired by the ability of humans to find salient characteristics in intricate views \cite{vaswani2017attention,guo2022attention,woo2018cbam}. Attention "implies the withdrawal from some things to deal effectively with others" and makes humans perceive, comprehend, and distinguish more effectively \cite{james2007principles}.
This paper discusses a new hybrid architecture that combines multiple CNNs, each leveraging attention to different color spaces.
The corresponding embeddings are then integrated by feature-level fusion \cite{qin2020ffa,dai2021attentional,10458131,10684224}.
Incorporating various color spaces in the PAD design is motivated by excellent results obtained in previous research \cite{marasco2022late,miao2023pa,zerman2019colornet,gowda2019colornet}. 

The contribution of this paper is summarized as follows:
\begin{itemize}
    \item  We propose the ColFigPhotoAttnNet architecture that combines multiple CNNs, each leveraging attention to different color spaces. This design aims to achieve robust finger photo PAD.
    
    \item Our study is the first to systematically analyze the performance of existing SOTA deep learning PAD systems, including convolutional and transformer-based models, in cross-capture device settings. To assess their generalization and robustness, we evaluated the proposed and existing methods using various capture devices, including iPhone 13 Pro (iPhone), Google Pixel 3 (google), Nokia C5 (Nokia), and OnePlus One (OPO).

    \item We extensively evaluate the performance of PAD systems in both \textit{inter-capture} and \textit{intra-capture} device scenarios.
\end{itemize}


The rest of the paper is organized as follows: Section 2 describes the review of the literature, Section 3 discusses the proposed approach, Section 4 presents our experimental results, and finally Section 5 draws conclusions. 

\section{Literature Review}
In this literature, we are only focusing on finger photo PAD research. Early studies explored PAD using handcrafted features, such as reflection properties, texture-based features like micro-textures, gradients, and light reflections, often classified using machine learning algorithms, particularly Support Vector Machines (SVMs) \cite{guo2010completed,kannala2012bsif,lowe1999object,hearst1998support}. 


In 2022, Marasco \textit{et al.}  introduced a framework that segments the finger region based on the distal phalange and converts it into multiple color spaces. Local patches around minutiae points are extracted and fed into an ensemble of pre-trained CNN architectures. Deeply learned features from these CNNs are combined using late fusion to make a global decision for distinguishing live versus spoof finger photo presentations. This method leverages texture patterns and minutiae information across diverse color representations for robust presentation attack detection \cite{marasco2022late}.

In 2023, Li \textit{et al.} conducted a comparative evaluation involving deep learning models like DenseNet 021, ResNet50, Efficient Net, Vision transformers, etc. Their study aimed to identify the most robust solution against this specific type of spoof attack \cite{li2023deep_g}. Also, in the same year, Li \textit{et al.} performed a detailed survey on deep learning techniques employed on finger photo PAD, showing that state-of-the-art methods have employed popular architectures like DenseNet variants, MobileNets-V2, MobileNet-V3, Efficient Net variants, and Vision Transformers \cite{li2023deep}. As PAD models, Purnapatra \textit{et al.} fine-tuned two deep learning architectures, DenseNet and NasNetMobile. By adapting the pre-trained weights of these models to the finger photo PAD task, they achieved impressive results with an Attack Presentation Classification Error Rate (APCER) of 0.00\%, Bona fide Presentation Classification Error Rate (BPCER) of 0.18\% against the known attack types in their dataset \cite{purnapatra2023presentation}. The authors explored incorporating color during the presentation attack instrument (PAI) casting process to obtain skin-colored PAIs. Adami \textit{et al.} proposed an unsupervised finger PAD method combining an autoencoder and convolutional block attention. This unsupervised approach achieved a BPCER of 0.96\% and APCER of 1.6\% \cite{adami2023contactless}.

In 2024, the researchers benchmarked the detection performance of eight popular pre-trained deep neural network models across three different finger photo segmentation schemes on a public finger photo dataset with four different presentation attack instruments \cite{li2024does}. The researchers proposed an explainable PAD framework using Grad-CAM and SNR metrics to quantify key features for detecting finger photo presentation attacks \cite{vurity2024}.

CNNs represent the prevailing architectural paradigm in computer vision, including architectures of MobileNet family \cite{mobilenetv2,howard2019searching}, VGG \cite{simonyan2014very}, ResNet \cite{he2016deep}, and EfficientNet \cite{tan2019efficientnet}.
More recently, different variants of Vision Transformers leveraging various attention mechanisms, 
exhibited superior performance in multiple tasks, including classification, segmentation, and detection, surpassing the capabilities of traditional CNNs \cite{dosovitskiy2020image,liu2021swin,touvron2021training,chen2021crossvit,park2024grafting,dong2022cswin}. 
For mobile devices, approaches such as MobileViT-XS, MobileViT-XXS, and Swin-Tiny have been introduced with different attention mechanisms to work on mobile platforms \cite{liu2021swin,mehta2021mobilevit}.


\section{Proposed Approach}
\begin{figure*}[t]
\centering
\begin{center}
\includegraphics[width= 17cm, height = 7.4 cm]{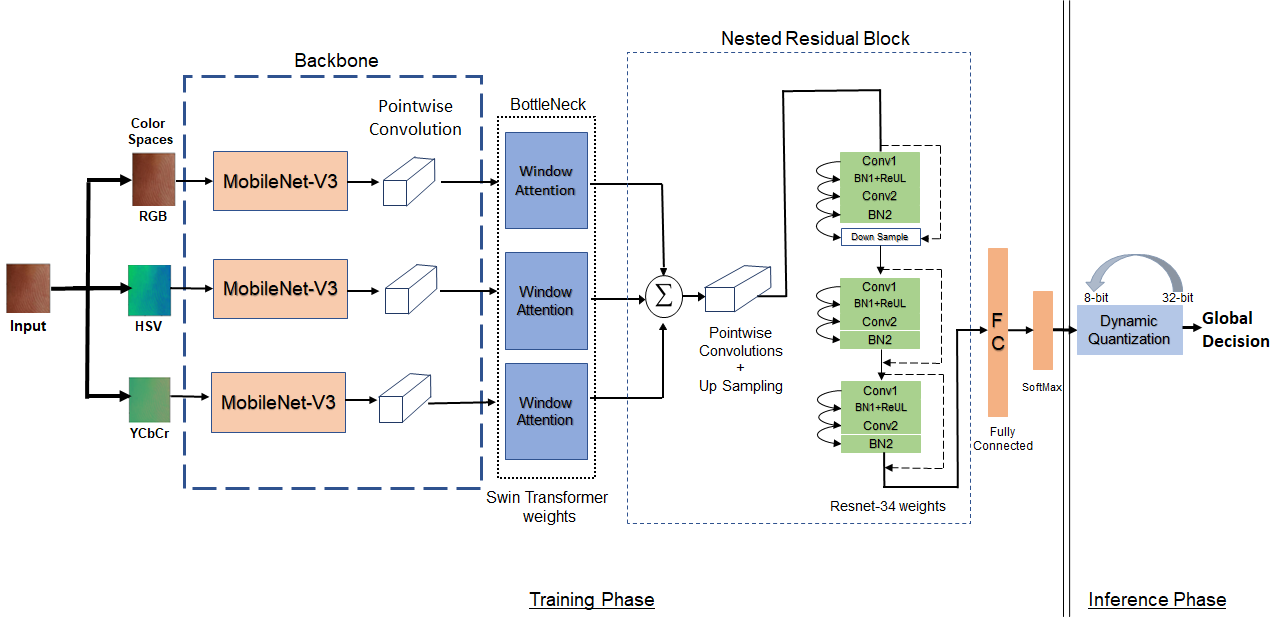}
\end{center}
   \caption{The architecture integrates MobileNet-V3 for feature extraction and applies pointwise convolution within a bottleneck framework with window attention mechanisms using fine-tuned Swin transformer weights. Then, features of three color spaces are combined with element-wise addition and pointwise convolution and fed in a Nested Residual Block that has been initialized with ResNet34 weights. Finally, at inference, the model applies Dynamic Quantization and gives the final global decision.}
\label{app}
\end{figure*}

A diagram for the proposed PAD system \textit{ColFigPhotoAttnNet} is illustrated in Figure \ref{app}.
The process starts by isolating the region of interest (ROI) and cropping the finger photo foreground using Faster R-CNN \cite{ren2016faster}. The input ROI is then converted to RGB, HSV, and YCbCr color space. It has been indicated by earlier studies that fusing multiple color spaces can improve finger photo PAD performance \cite{marasco2022late}. Hence, this work leverages different color models which possess complementary information. The backbone consists of three distinct MobileNet V3 Large networks that process each color space separately. These models have a lightweight architecture, thus suitable for mobile applications with limited computational resources.
The architecture is initialized using pre-trained weights from models previously trained on ImageNet \cite{imagenet}. Features extracted by individual networks are subjected to pointwise convolution followed by a bottleneck framework where each network incorporates a window attention mechanism. The weights of the attention layer are initialized with predefined weights of the Swin transformer fine-tuned on finger photos.

The features from the three attention layers, representing various color spaces, are combined via element-wise addition, mixing channels through pointwise convolution. The output is fed into a Nested Residual Block, which consists of convolutional layers with skip connections.  
The blocks are initially set with weights from a previously trained ResNet-34 model. 
Finally, a fully connected layer with a SoftMax layer leads to a global decision. Dynamic quantization (DQ) is applied to compact the model's size and enhance deployment efficiency on various mobile platforms. 

\subsection{Window Attention on Color Spaces}


Window attention is a variant of the attention mechanism that considers local regions (windows) in the input feature map.
The input color feature map in the proposed framework is partitioned into non-overlapping tiles, each being a 7x7 sub-region. Within each of these 7x7 tiles, self-attention is computed independently, allowing the model to capture relationships and dependencies within localized regions of the input feature map. The window attention mechanism takes as input feature maps represented in $d = 768$ dimensions. This
mechanism partitions inputs non-overlapping windows where each has height $Wh = Ww = 7$, width denoted by $Ww$, and height denoted by $Wh$ respectively. Consequently, a single window comprises $N = W_h \times W_w = 49$ tokens. To efficiently manage high-dimensional input within these localized regions, the mechanism employs $h = 24$ attention heads, each operating on a reduced dimension of $d_h = d / h = 32$. This setup facilitates parallel processing across multiple aspects of the input within each window, enhancing the model's capability to distill pertinent features from localized segments of the feature map.

The $Q, K, V$ indicate the $query, key, value$ matrices within each window and per head, and are generated by applying a linear layer on $X\in \mathbb{R}^{N \times d_{h}}$, the feature map of each window, respectively.

\begin{equation}
\small{
Q, K, V = \text{Linear}(X) \in \mathbb{R}^{N \times d_{h}}
}
\end{equation}

The window attention in each head can be written as follows:

\begin{equation}
\small{
Attention(Q, K, V) = \text{SoftMax}\left(\frac{QK^T}{\sqrt{d_h}}+B\right)V
}
\end{equation}

where $B \in \mathbb{R}^{N \times N}$ is a relative position bias per head. The output features of the attention operation between $Q$, $K$, and $V$ for multiple heads are mixed through a linear layer.

\subsection{Nested Residual Block}
The Nested Residual Block is a composite structure within a neural network that enhances input features $X$ using convolutional layers, normalization techniques, and non-linear activations integrated into a residual learning framework. This block initially transforms through a convolutional layer $Conv1$ followed by batch normalization $BN_1$ and activation of the Rectified Linear Unit (ReLU).

\begin{equation}
\small{
X_1 = \text{ReLU}(BN_1(Conv_1(X)))
}
\end{equation}

Then, it applies downsampling $X_{\text{down}}$ to extract higher-level features and further process them. Subsequently, the samplers are re-sampled to their dimensions $X_{\text{up}}$. The block culminates by merging the $X_{\text{up}}$ features with the initial transformation via a residual connection $X_{\text{res}}$ and produces the final output $Y$ through another convolution $Conv2$ and batch normalization $BN2$.


\begin{equation}
\small{
Y = BN_2(Conv_2(X_{\text{res}}))
}
\end{equation}

This design enables the network to capture information effectively on multiple scales while ensuring a smooth gradient flow for training models.

\subsection{8 bit- Dynamic Quantization}
The dynamic quantization (DQ) mechanism concentrates more on the weights during inference, helping to compress the most memory-consuming sections of the model efficiently. This is a dynamic process since it does scale and zero-point calculations at runtime that are adaptive to the actual data distribution; thus, accuracy is maintained regardless of decreased precision. To optimize for model tensor and inference speed, this method maps 8-bit integers in the range \([-128, 127]\) by floating point values representation.

\begin{algorithm}
\small{
\caption{Dynamic Quantization}
\begin{algorithmic}[1]
\State Identify $F_{\min}$ and $F_{\max}$, the minimum and maximum values in tensor $F$ from a model.
\State Calculate scale $S$ as $\frac{F_{\max} - F_{\min}}{255}$.
\State Calculate zero point $Z$ as $\text{round}\left(\frac{-F_{\min}}{S}\right) - 128$.
\For{each value $f$ in $F$}
    \State Quantize $f$ to $Q = \text{round}\left(\frac{f - F_{\min}}{S}\right) - 128$.
\EndFor
\For{each value $q$ in $Q$}
    \State Dequantize $q$ to $F' = (q + 128) \times S + F_{\min}$.
\EndFor
\State \Return Quantized tensor $Q$ and dequantized tensor $F'$.
\end{algorithmic}
}
\end{algorithm}

Algorithm 1 provides an overview of how dynamic quantization works – it converts floating point tensors into 8-bit ones by scaling them down concerning their maximum absolute value before rounding off. By quantizing model weights as lower-precision integers, we can decrease the model's memory usage and computational demands while limiting its parameter size. Such compression methods are beneficial for deploying deep networks on smartphones or other resource-limited hardware without sacrificing much accuracy \cite{gholami2022survey}.

\section{Experimental Results}
In this study, we meticulously evaluated the proposed framework, initially focusing on \textit{inter-capture} device variations, followed by \textit{intra-capture} device variations, and concluding with a structured ablation study. This comprehensive approach allowed us to analyze the contribution of individual components to overall performance and rigorously validate the model's robustness across various capture scenarios. In Section 4.1, the particular databases used in this work are explained in more detail. Section 4.2 outlines the specifics of the training and evaluation technique used. The research results, including an \textit{inter-capture}, \textit{intra-capture} device, and ablation study, are presented in Section 4.3.

\subsection{Dataset}
In this work, we are using three datasets to evaluate the performance of various models on finger photo PAD. 

\textbf{Database \#1:} This dataset includes images of 112 subjects captured with an iPhone 13 Pro, both in bona fide and spoofed finger photos in various conditions, both indoors and outdoors. Each subject contributed 24 images, adding up to 11,648 spoofed images. The spoofed images have display and printout attacks using several devices, such as the Samsung Tab 7+,
iPad Pro and others represent a wide range of attack scenarios \cite{MFPADdatabase}.

\textbf{Database \#2:} This dataset comprises 40 images from 100 subjects, both live and spoof finger photos. The images were captured using the Google Pixel 3 XL camera, with particular attention to various indoor and outdoor lighting conditions and subject demographics. The spoofed portion of the database includes 20,800 images fabricated using multiple devices, designed to test the system's resilience against various spoofing mechanisms \cite{MFPADdatabase}.

\textbf{Database \#3:} Originating from live captures using an iPhone 5 camera, this data set contains 64 subjects, which was extended to include spoofed images for research of photo and print attack mechanisms. The live database contains 4,096 images categorized into groups based on subjects, fingers, variations, and samples. For spoofing, OnePlusOne and Nokia devices are used as capture mechanisms to study different spoofs by displaying images on devices like an iPad and Dell Inspiron, as well as high-quality printouts. There are 8192 spoof images and live finger photos \cite{taneja}.

\subsection{Evaluation Protocol}
The models, originally trained on the ImageNet dataset, were carefully fine-tuned. We maintained a consistent batch size of 32 and 20 epochs throughout the training. The training set was organized based on each experiment's capture device, including the iPhone, Google, OPO, and Nokia, considering only indoor lighting conditions. The experiments were conducted on finger patches, which were extracted using the Faster R-CNN algorithm, which was trained on finger photos \cite{marasco2021fingerphoto}. The images in which R-CNN failed to extract ROI were removed. The composition of the remaining data set is as follows:

The iPhone 13 Pro database contains 2,688 bona fide and 5,718 Attacks. The Google dataset includes 1,955 bona fide and 9,141 Attacks, while the OnePlusOne has 2,048 bona fide and 1,973 Attacks. Lastly, the Nokia dataset comprises 2,048 bona fide and 2,043 Attacks. The data distribution follows a 50\% training, 30\% validation, and 20\% testing split, with subjects mutually exclusive.

We use the performance metrics defined by the International Organization for Standardization (ISO/IEC SC 37): Attack presentation classification error rate (APCER), bona fide presentation classification error rate (BPCER), and equal error rate (EER) \cite{nist}. APCER measures the proportion of attack attempts (PAs), which the system mistakenly identifies as Bona fide (live). In contrast, BPCER measures the proportion of bona fide attempts wrongly classified as attacks. EER\% is a metric used to evaluate the performance of biometric systems. Indicates the point at which APCER and BPCER are equal.
The results show BPCER\% when APCER is 5\%  and 10\% , respectively.


\subsection{Results}

\begin{table}[]
\small
\renewcommand{\arraystretch}{1}
\setlength{\tabcolsep}{0.9pt}
\footnotesize
\begin{tabular}{|c||cc||cc||cc||cc|}
\hline
\multirow{2}{*}{\textbf{\begin{tabular}[c]{@{}c@{}} Baseline \end{tabular}}} & \multicolumn{2}{c|}{\textbf{Google}} & \multicolumn{2}{c|}{\textbf{iPhone}} & \multicolumn{2}{c|}{\textbf{OPO}} & \multicolumn{2}{c|}{\textbf{Nokia}} \\ \cline{2-9} 
 & \multicolumn{1}{c|}{\textbf{5\%}} & \textbf{10\%} & \multicolumn{1}{c|}{\textbf{5\%}} & \textbf{10\%} & \multicolumn{1}{c|}{\textbf{5\%}} & \textbf{10\%} & \multicolumn{1}{c|}{\textbf{5\%}} & \textbf{10\%} \\ \hline
MobileNet-V3 L \cite{MFPADdatabase} & \multicolumn{1}{c|}{35.03} & 14.01 & \multicolumn{1}{c|}{34.51} & 23.22 & \multicolumn{1}{c|}{8.05} & 0.57 & \multicolumn{1}{c|}{8.33} & 1.56 \\ \hline
MobileNet-V3 S \cite{marasco2021deep} & \multicolumn{1}{c|}{37.34} & 18.84 & \multicolumn{1}{c|}{36.46} & 17.96 & \multicolumn{1}{c|}{1.59} & 0.53 & \multicolumn{1}{c|}{2.12} & 0.53 \\ \hline
Mobile VIT XS \cite{li2023deep_g} & \multicolumn{1}{c|}{36.57} & 23.17 & \multicolumn{1}{c|}{62.97} & 50.14 & \multicolumn{1}{c|}{21.2} & 10.87 & \multicolumn{1}{c|}{28.64} & 21.11 \\ \hline
Mobile VIT XXS \cite{li2023deep_g} & \multicolumn{1}{c|}{34.19} & 25.14 & \multicolumn{1}{c|}{53} & 40.55 & \multicolumn{1}{c|}{21.31} & 15.3 & \multicolumn{1}{c|}{20.32} & 18.18 \\ \hline
EfficientNet-B7 \cite{li2024does} & \multicolumn{1}{c|}{28.07} & 13.72 & \multicolumn{1}{c|}{26.81} & 17.39 & \multicolumn{1}{c|}{1.22} & 0.52 & \multicolumn{1}{c|}{4.84} & 2.15 \\ \hline
EfficientNet-B5 \cite{li2024does} & \multicolumn{1}{c|}{27.11} & 14.81 & \multicolumn{1}{c|}{27.11} & 18.01 & \multicolumn{1}{c|}{2.56} & 1.03 & \multicolumn{1}{c|}{13.54} & 4.08 \\ \hline
Swin Transformer \cite{adami2023contactless} & \multicolumn{1}{c|}{21.98} & 15.93 & \multicolumn{1}{c|}{39.98} & 29.96 & \multicolumn{1}{c|}{1.64} & 0.27 & \multicolumn{1}{c|}{1.05} & 0.52 \\ \hline
\textbf{ColFigPhotoAttnNet} & \multicolumn{1}{c|}{\textbf{20.24}} & \textbf{6.73} & \multicolumn{1}{c|}{\textbf{12.78}} & \textbf{6.21} & \multicolumn{1}{c|}{\textbf{1.11}} & \textbf{0.21} & \multicolumn{1}{c|}{\textbf{0.11}} & \textbf{0.19} \\ \hline
\end{tabular}
\caption{The table shows \textit{intra-capture} device performance using BPCER\% @ APCER= 5\% and APCER= 10\%.} 

\label{intra}

\end{table}

\begin{table*}[]
\begin{center}
\small
\setlength{\tabcolsep}{1.2 pt} 
\renewcommand{\arraystretch}{1} 

\begin{tabular}{|c|c||cc||cc||cc||cc||cl|}
\hline
\multirow{3}{*}{\textbf{\begin{tabular}[c]{@{}c@{}} Baseline \end{tabular}}} & \multirow{3}{*}{\textbf{Parameters}} & \multicolumn{8}{c||}{\textbf{Training}} & \multicolumn{2}{c|}{\multirow{3}{*}{\textbf{Testing}}} \\ \cline{3-10}
 & & \multicolumn{2}{c||}{\textbf{Google}} & \multicolumn{2}{c||}{\textbf{iPhone}} & \multicolumn{2}{c||}{\textbf{OPO}} & \multicolumn{2}{c||}{\textbf{Nokia}} & \multicolumn{2}{c|}{} \\ \cline{3-10}
 & & \multicolumn{1}{c|}{\textbf{5\%}} & \multicolumn{1}{c||}{\textbf{10\%}} & \multicolumn{1}{c|}{\textbf{5\%}} & \multicolumn{1}{c||}{\textbf{10\%}} & \multicolumn{1}{c|}{\textbf{5\%}} & \multicolumn{1}{c||}{\textbf{10\%}} & \multicolumn{1}{c|}{\textbf{5\%}} & \multicolumn{1}{c||}{\textbf{10\%}} & \multicolumn{2}{c|}{} \\ \hline
\multirow{4}{*}{MobileNet V3 Large \cite{MFPADdatabase}} & \multirow{4}{*}{5.4 M} & \multicolumn{1}{c|}{\textbf{-}} & \multicolumn{1}{c||}{\textbf{-}} & \multicolumn{1}{c|}{\textbf{57.48}} & \multicolumn{1}{c||}{\textbf{45.36}} & \multicolumn{1}{c|}{98.9} & \multicolumn{1}{c||}{93.37} & \multicolumn{1}{c|}{98.96} & \multicolumn{1}{c||}{96.37} & \multicolumn{2}{c|}{\textbf{Google}} \\ \cline{3-12} 
 & & \multicolumn{1}{c|}{67.72} & \multicolumn{1}{c||}{53.43} & \multicolumn{1}{c|}{\textbf{-}} & \multicolumn{1}{c||}{\textbf{-}} & \multicolumn{1}{c|}{100} & \multicolumn{1}{c||}{100} & \multicolumn{1}{c|}{99} & \multicolumn{1}{c||}{98.5} & \multicolumn{2}{c|}{\textbf{iPhone}} \\ \cline{3-12} 
 & & \multicolumn{1}{c|}{95.8} & \multicolumn{1}{c||}{90.91} & \multicolumn{1}{c|}{98.39} & \multicolumn{1}{c||}{95.98} & \multicolumn{1}{c|}{\textbf{-}} & \multicolumn{1}{c||}{\textbf{-}} & \multicolumn{1}{c|}{19.79} & \multicolumn{1}{c||}{7.49} & \multicolumn{2}{c|}{\textbf{OPO}} \\ \cline{3-12} 
 & & \multicolumn{1}{c|}{96.45} & \multicolumn{1}{c||}{90.91} & \multicolumn{1}{c|}{96.39} & \multicolumn{1}{c||}{94.49} & \multicolumn{1}{c|}{16.29} & \multicolumn{1}{c||}{9.55} & \multicolumn{1}{c|}{\textbf{-}} & \multicolumn{1}{c||}{\textbf{-}} & \multicolumn{2}{c|}{\textbf{Nokia}} \\ \hline
\multirow{4}{*}{MobileNet V3 Small \cite{marasco2021deep}} & \multirow{4}{*}{2.5 M} & \multicolumn{1}{c|}{\textbf{-}} & \multicolumn{1}{c||}{\textbf{-}} & \multicolumn{1}{c|}{65.62} & \multicolumn{1}{c||}{50.32} & \multicolumn{1}{c|}{96.09} & \multicolumn{1}{c||}{92.18} & \multicolumn{1}{c|}{95.48} & \multicolumn{1}{c||}{99.85} & \multicolumn{2}{c|}{\textbf{Google}} \\ \cline{3-12} 
 & & \multicolumn{1}{c|}{50} & \multicolumn{1}{c||}{48.2} & \multicolumn{1}{c|}{\textbf{-}} & \multicolumn{1}{c||}{\textbf{-}} & \multicolumn{1}{c|}{97.89} & \multicolumn{1}{c||}{96.32} & \multicolumn{1}{c|}{99.48} & \multicolumn{1}{c||}{96.37} & \multicolumn{2}{c|}{\textbf{iPhone}} \\ \cline{3-12} 
 & & \multicolumn{1}{c|}{92.91} & \multicolumn{1}{c||}{88.13} & \multicolumn{1}{c|}{98.48} & \multicolumn{1}{c||}{96.96} & \multicolumn{1}{c|}{\textbf{-}} & \multicolumn{1}{c||}{\textbf{-}} & \multicolumn{1}{c|}{11.05} & \multicolumn{1}{c||}{3.16} & \multicolumn{2}{c|}{\textbf{OPO}} \\ \cline{3-12} 
 & & \multicolumn{1}{c|}{86.27} & \multicolumn{1}{c||}{77.28} & \multicolumn{1}{c|}{97.27} & \multicolumn{1}{c||}{95.17} & \multicolumn{1}{c|}{12.5} & \multicolumn{1}{c||}{\textbf{3.12}} & \multicolumn{1}{c|}{\textbf{-}} & \multicolumn{1}{c||}{\textbf{-}} & \multicolumn{2}{c|}{\textbf{Nokia}} \\ \hline
\multirow{4}{*}{Mobile VIT-XS  \cite{li2023deep_g}} & \multirow{4}{*}{2.3 M} & \multicolumn{1}{c|}{\textbf{-}} & \multicolumn{1}{c||}{\textbf{-}} & \multicolumn{1}{c|}{84.01} & \multicolumn{1}{c||}{74.33} & \multicolumn{1}{c|}{89.13} & \multicolumn{1}{c||}{85.33} & \multicolumn{1}{c|}{97.38} & \multicolumn{1}{c||}{93.19} & \multicolumn{2}{c|}{\textbf{Google}} \\ \cline{3-12} 
 & & \multicolumn{1}{c|}{78.75} & \multicolumn{1}{c||}{62.51} & \multicolumn{1}{c|}{\textbf{-}} & \multicolumn{1}{c||}{\textbf{-}} & \multicolumn{1}{c|}{91.88} & \multicolumn{1}{c||}{84.26} & \multicolumn{1}{c|}{97.75} & \multicolumn{1}{c||}{93.82} & \multicolumn{2}{c|}{\textbf{iPhone}} \\ \cline{3-12} 
 & & \multicolumn{1}{c|}{70.57} & \multicolumn{1}{c||}{64.47} & \multicolumn{1}{c|}{90.2} & \multicolumn{1}{c||}{85} & \multicolumn{1}{c|}{\textbf{-}} & \multicolumn{1}{c||}{\textbf{-}} & \multicolumn{1}{c|}{72.9} & \multicolumn{1}{c||}{62.18} & \multicolumn{2}{c|}{\textbf{OPO}} \\ \cline{3-12} 
 & & \multicolumn{1}{c|}{70.64} & \multicolumn{1}{c||}{60.77} & \multicolumn{1}{c|}{95.22} & \multicolumn{1}{c||}{92.43} & \multicolumn{1}{c|}{87.43} & \multicolumn{1}{c||}{82.2} & \multicolumn{1}{c|}{\textbf{-}} & \multicolumn{1}{c||}{\textbf{-}} & \multicolumn{2}{c|}{\textbf{Nokia}} \\ \hline
\multirow{4}{*}{Mobile VIT-XXS \cite{li2023deep_g}} & \multirow{4}{*}{1.3 M} & \multicolumn{1}{c|}{\textbf{-}} & \multicolumn{1}{c||}{\textbf{-}} & \multicolumn{1}{c|}{83.85} & \multicolumn{1}{c||}{75.51} & \multicolumn{1}{c|}{84.54} & \multicolumn{1}{c||}{79.38} & \multicolumn{1}{c|}{97.96} & \multicolumn{1}{c||}{84.69} & \multicolumn{2}{c|}{\textbf{Google}} \\ \cline{3-12} 
 & & \multicolumn{1}{c|}{55.32} & \multicolumn{1}{c||}{36} & \multicolumn{1}{c|}{\textbf{-}} & \multicolumn{1}{c||}{\textbf{-}} & \multicolumn{1}{c|}{96.26} & \multicolumn{1}{c||}{93.58} & \multicolumn{1}{c|}{97.09} & \multicolumn{1}{c||}{92.72} & \multicolumn{2}{c|}{\textbf{iPhone}} \\ \cline{3-12} 
 & & \multicolumn{1}{c|}{52.17} & \multicolumn{1}{c||}{42.98} & \multicolumn{1}{c|}{88.58} & \multicolumn{1}{c||}{82.14} & \multicolumn{1}{c|}{\textbf{-}} & \multicolumn{1}{c||}{\textbf{-}} & \multicolumn{1}{c|}{41.11} & \multicolumn{1}{c||}{35.56} & \multicolumn{2}{c|}{\textbf{OPO}} \\ \cline{3-12} 
 & & \multicolumn{1}{c|}{57.44} & \multicolumn{1}{c||}{40.75} & \multicolumn{1}{c|}{93.93} & \multicolumn{1}{c||}{85.85} & \multicolumn{1}{c|}{58.01} & \multicolumn{1}{c||}{49.17} & \multicolumn{1}{c|}{\textbf{-}} & \multicolumn{1}{c||}{\textbf{-}} & \multicolumn{2}{c|}{\textbf{Nokia}} \\ \hline
\multirow{4}{*}{EfficientNet-B7 \cite{li2024does}} & \multirow{4}{*}{66.34 M} & \multicolumn{1}{c|}{\textbf{-}} & \multicolumn{1}{c||}{\textbf{-}} & \multicolumn{1}{c|}{65.87} & \multicolumn{1}{c||}{48.99} & \multicolumn{1}{c|}{93.96} & \multicolumn{1}{c||}{86.81} & \multicolumn{1}{c|}{95.98} & \multicolumn{1}{c||}{90.45} & \multicolumn{2}{c|}{\textbf{Google}} \\ \cline{3-12} 
 & & \multicolumn{1}{c|}{70.92} & \multicolumn{1}{c||}{58.54} & \multicolumn{1}{c|}{\textbf{-}} & \multicolumn{1}{c||}{\textbf{-}} & \multicolumn{1}{c|}{98.42} & \multicolumn{1}{c||}{93.68} & \multicolumn{1}{c|}{99.47} & \multicolumn{1}{c||}{98.41} & \multicolumn{2}{c|}{\textbf{iPhone}} \\ \cline{3-12} 
 & & \multicolumn{1}{c|}{97.82} & \multicolumn{1}{c||}{95.83} & \multicolumn{1}{c|}{88.96} & \multicolumn{1}{c||}{82.22} & \multicolumn{1}{c|}{\textbf{-}} & \multicolumn{1}{c||}{\textbf{-}} & \multicolumn{1}{c|}{21.11} & \multicolumn{1}{c||}{9.55} & \multicolumn{2}{c|}{\textbf{OPO}} \\ \cline{3-12} 
 & & \multicolumn{1}{c|}{96.6} & \multicolumn{1}{c||}{92.73} & \multicolumn{1}{c|}{82.77} & \multicolumn{1}{c||}{74.16} & \multicolumn{1}{c|}{\textbf{7.03}} & \multicolumn{1}{c||}{6.49} & \multicolumn{1}{c|}{\textbf{-}} & \multicolumn{1}{c||}{\textbf{-}} & \multicolumn{2}{c|}{\textbf{Nokia}} \\ \hline
\multirow{4}{*}{EfficientNet-B5 \cite{li2024does}} & \multirow{4}{*}{27.89 M} & \multicolumn{1}{c|}{\textbf{-}} & \multicolumn{1}{c||}{\textbf{-}} & \multicolumn{1}{c|}{60.44} & \multicolumn{1}{c||}{48.04} & \multicolumn{1}{c|}{99.46} & \multicolumn{1}{c||}{97.83} & \multicolumn{1}{c|}{95.41} & \multicolumn{1}{c||}{90.82} & \multicolumn{2}{c|}{\textbf{Google}} \\ \cline{3-12} 
 & & \multicolumn{1}{c|}{67.75} & \multicolumn{1}{c||}{57.33} & \multicolumn{1}{c|}{\textbf{-}} & \multicolumn{1}{c||}{\textbf{-}} & \multicolumn{1}{c|}{99.52} & \multicolumn{1}{c||}{99.03} & \multicolumn{1}{c|}{100} & \multicolumn{1}{c||}{99.48} & \multicolumn{2}{c|}{\textbf{iPhone}} \\ \cline{3-12} 
 & & \multicolumn{1}{c|}{96.9} & \multicolumn{1}{c||}{94.98} & \multicolumn{1}{c|}{89.84} & \multicolumn{1}{c||}{84.23} & \multicolumn{1}{c|}{\textbf{-}} & \multicolumn{1}{c||}{\textbf{-}} & \multicolumn{1}{c|}{31.66} & \multicolumn{1}{c||}{18.09} & \multicolumn{2}{c|}{\textbf{OPO}} \\ \cline{3-12} 
 & & \multicolumn{1}{c|}{88.08} & \multicolumn{1}{c||}{81.84} & \multicolumn{1}{c|}{94.27} & \multicolumn{1}{c||}{91.46} & \multicolumn{1}{c|}{9.49} & \multicolumn{1}{c||}{3.31} & \multicolumn{1}{c|}{\textbf{-}} & \multicolumn{1}{c||}{\textbf{-}} & \multicolumn{2}{c|}{\textbf{Nokia}} \\ \hline
\multirow{4}{*}{Swin Transformer \cite{adami2023contactless}} & \multirow{4}{*}{30.38 M} & \multicolumn{1}{c|}{\textbf{-}} & \multicolumn{1}{c||}{\textbf{-}} & \multicolumn{1}{c|}{78.02} & \multicolumn{1}{c||}{66.94} & \multicolumn{1}{c|}{97.21} & \multicolumn{1}{c||}{95.53} & \multicolumn{1}{c|}{96.84} & \multicolumn{1}{c||}{94.21} & \multicolumn{2}{c|}{\textbf{Google}} \\ \cline{3-12} 
 & & \multicolumn{1}{c|}{51.91} & \multicolumn{1}{c||}{42.21} & \multicolumn{1}{c|}{\textbf{-}} & \multicolumn{1}{c||}{\textbf{-}} & \multicolumn{1}{c|}{99.44} & \multicolumn{1}{c||}{98.87} & \multicolumn{1}{c|}{99.5} & \multicolumn{1}{c||}{99.5} & \multicolumn{2}{c|}{\textbf{iPhone}} \\ \cline{3-12} 
 & & \multicolumn{1}{c|}{95.16} & \multicolumn{1}{c||}{92.2} & \multicolumn{1}{c|}{88.61} & \multicolumn{1}{c||}{73.86} & \multicolumn{1}{c|}{\textbf{-}} & \multicolumn{1}{c||}{\textbf{-}} & \multicolumn{1}{c|}{11.64} & \multicolumn{1}{c||}{4.76} & \multicolumn{2}{c|}{\textbf{OPO}} \\ \cline{3-12} 
 & & \multicolumn{1}{c|}{94.79} & \multicolumn{1}{c||}{90.4} & \multicolumn{1}{c|}{91.07} & \multicolumn{1}{c||}{86.64} & \multicolumn{1}{c|}{26.42} & \multicolumn{1}{c||}{\textbf{3.63}} & \multicolumn{1}{c|}{\textbf{-}} & \multicolumn{1}{c||}{\textbf{-}} & \multicolumn{2}{c|}{\textbf{Nokia}} \\ \hline
\multirow{4}{*}{ColFigPhotoAttnNet (Our)} & \multirow{4}{*}{24.89 M} & \multicolumn{1}{c|}{\textbf{-}} & \multicolumn{1}{c||}{\textbf{-}} & \multicolumn{1}{c|}{\textbf{65.78}} & \multicolumn{1}{c||}{\textbf{51.84}} & \multicolumn{1}{c|}{86.24} & \multicolumn{1}{c||}{51.84} & \multicolumn{1}{c|}{89.23} & \multicolumn{1}{c||}{92.59} & \multicolumn{2}{c|}{\textbf{Google}} \\ \cline{3-12} 
 & & \multicolumn{1}{c|}{\textbf{37.96}} & \multicolumn{1}{c||}{\textbf{26.09}} & \multicolumn{1}{c|}{\textbf{-}} & \multicolumn{1}{c||}{\textbf{-}} & \multicolumn{1}{c|}{93.65} & \multicolumn{1}{c||}{86.772} & \multicolumn{1}{c|}{92.82} & \multicolumn{1}{c||}{85.12} & \multicolumn{2}{c|}{\textbf{iPhone}} \\ \cline{3-12} 
 & & \multicolumn{1}{c|}{79.24} & \multicolumn{1}{c||}{73.337} & \multicolumn{1}{c|}{91.597} & \multicolumn{1}{c||}{85.22} & \multicolumn{1}{c|}{\textbf{-}} & \multicolumn{1}{c||}{\textbf{-}} & \multicolumn{1}{c|}{\textbf{3.092}} & \multicolumn{1}{c||}{\textbf{1.03}} & \multicolumn{2}{c|}{\textbf{OPO}} \\ \cline{3-12} 
 & & \multicolumn{1}{c|}{81.36} & \multicolumn{1}{c||}{68.79} & \multicolumn{1}{c|}{79.29} & \multicolumn{1}{c||}{66.573} & \multicolumn{1}{c|}{\textbf{11.22}} & \multicolumn{1}{c||}{\textbf{4.812}} & \multicolumn{1}{c|}{\textbf{-}} & \multicolumn{1}{c||}{\textbf{-}} & \multicolumn{2}{c|}{\textbf{Nokia}} \\ \hline
\end{tabular}
\end{center}

\caption{The table shows BPCER @ APCER 5\% and 10\% of different models in \textit{inter-capture} settings.}

\label{inter}

\end{table*}

\begin{table}[]
\begin{center}
\renewcommand{\arraystretch}{1.1}
\setlength{\tabcolsep}{2.5pt}
\footnotesize

\begin{tabular}{|ccc|c|c|c|c|c|}
\hline
\multicolumn{3}{|c|}{\textbf{Color Spaces}} & \textbf{BottleNeck} & \multirow{2}{*}{\textbf{\begin{tabular}[c]{@{}c@{}}Residual \\ Block\end{tabular}}} & \textbf{DQ} &\textbf{BPCER} & \textbf{BPCER} \\ \cline{1-4} \cline{7-8} 
\multicolumn{1}{|c|}{\textbf{RGB}} & \multicolumn{1}{c|}{\textbf{HSV}} & \textbf{YCbCr} & \textbf{Attention} &  &  & \textbf{5\%} & \textbf{10\%} \\ \hline
\multicolumn{1}{|c|}{\checkmark} & \multicolumn{1}{c|}{x} & x & \checkmark & \checkmark & x & 16.77 & 10.94 \\ \hline
\multicolumn{1}{|c|}{\checkmark} & \multicolumn{1}{c|}{\checkmark} & x & \checkmark & \checkmark & x & 14.12 & 7.868 \\ \hline
\multicolumn{1}{|c|}{\checkmark} & \multicolumn{1}{c|}{x} & \checkmark & \checkmark & \checkmark & x & 20.84 & 11.74 \\ \hline
\multicolumn{1}{|c|}{\checkmark} & \multicolumn{1}{c|}{\checkmark} & \checkmark & \checkmark & x & x & 21.79 & 10.71 \\ \hline
\multicolumn{1}{|c|}{\checkmark} & \multicolumn{1}{c|}{\checkmark} & \checkmark & x & \checkmark & x & 24.16 & 12.92 \\ \hline
\multicolumn{1}{|c|}{\checkmark} & \multicolumn{1}{c|}{\checkmark} & \checkmark & \checkmark & \checkmark & x & 9.32 & 5.03 \\ \hline
\multicolumn{1}{|c|}{\checkmark} & \multicolumn{1}{c|}{\checkmark} & \checkmark & \checkmark & \checkmark & \checkmark & 12.78 & 6.21 \\ \hline
\end{tabular}
\end{center}
\caption{Ablation study using Database 1, This table shows BPCER @APCER 5\% and APCER 10\%.} 
\label{ablation}
\end{table}

\begin{figure*}[htbp]
\centering
\begin{subfigure}{0.48\textwidth}
\centering
\includegraphics[width=\linewidth, height=0.28\textheight]{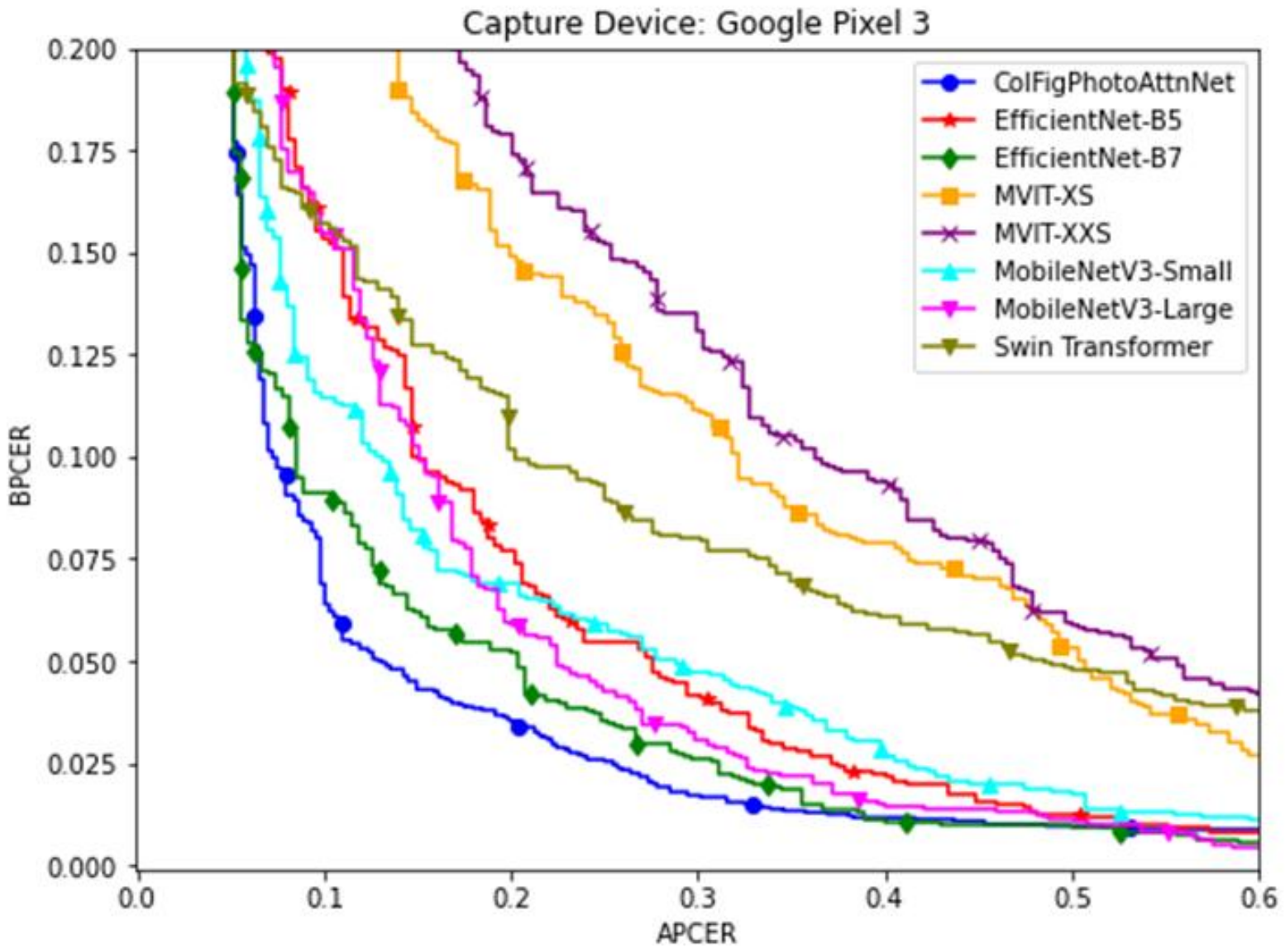}
\caption{Google}
\end{subfigure}
\hfill
\begin{subfigure}{0.48\textwidth}
\centering
\includegraphics[width=\linewidth, height=0.28\textheight]{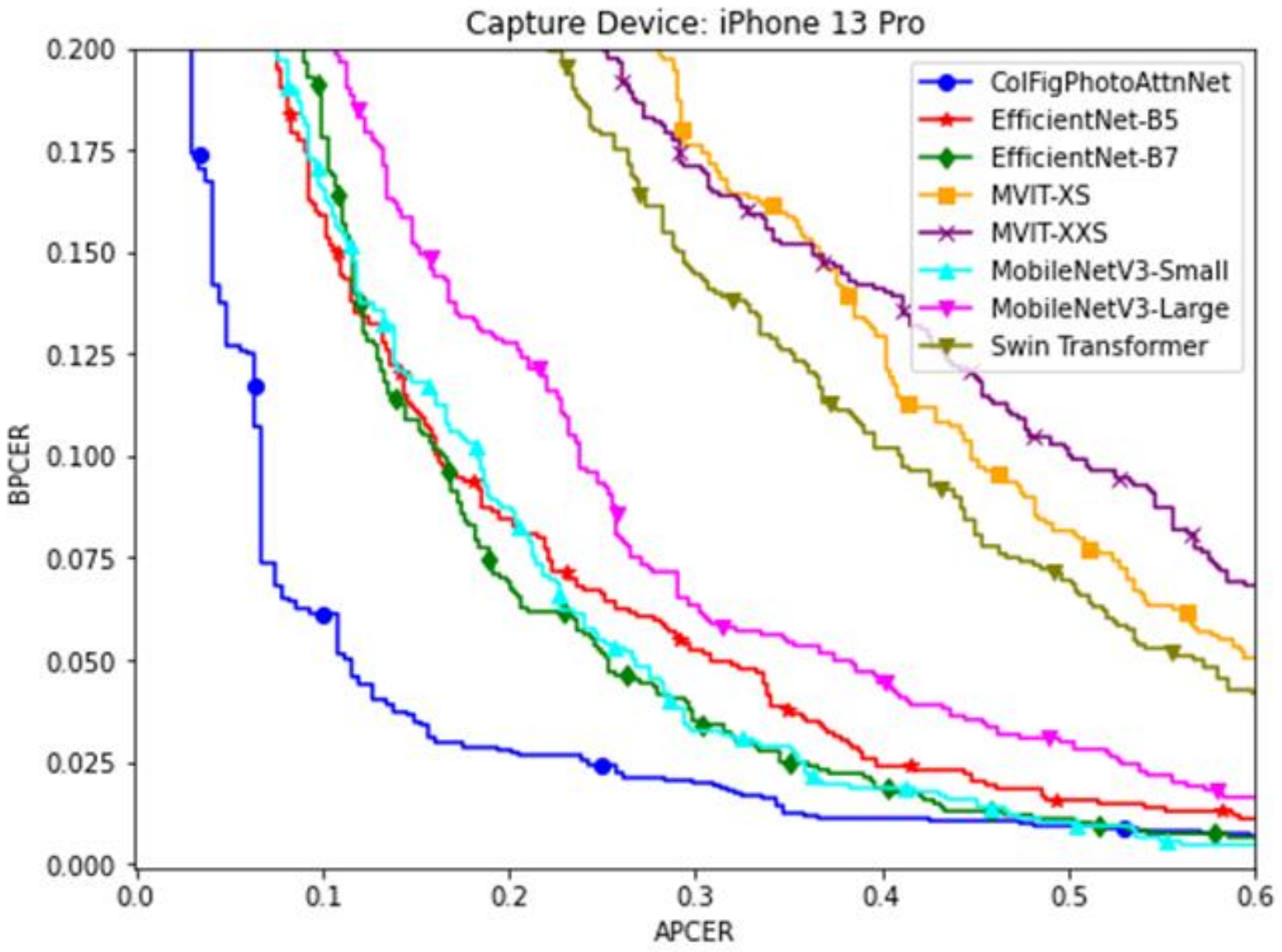}
\caption{iPhone}
\end{subfigure}

\caption{DET curves showing the performance of the models on iPhone 13 Pro and Google Pixel 3 capture devices}
\label{roc1}
\end{figure*}

\begin{figure}[h]
\centering
\includegraphics[width=8.2cm, height=6cm]{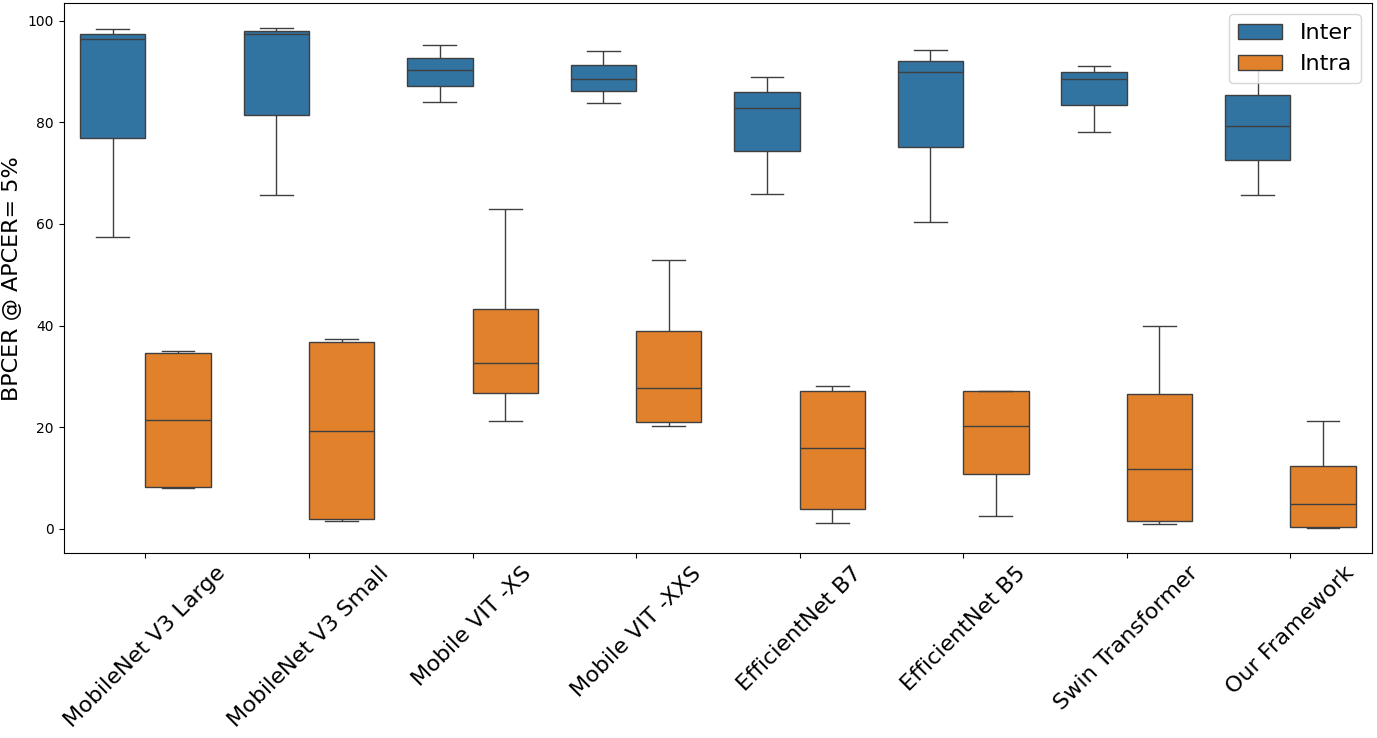}
\caption{Boxplot showing the interoperability of the models}
\label{boxplot}
\end{figure}

In this paper, we are utilizing the best finger photo PAD existing models as a baseline for comparison. We systematically compare the performance of our proposed framework with various deep networks, including MobileNet-V3 large and small, MobileViT XS and XXS, EfficientNet-B5, EfficientNet-B7, and Swin Transformer. The models are selected based on their parameter range from 1 Million to 66 Million to understand the trade-off between model deployment on mobile devices and the robustness of the PAD system. We conduct an ablation study to understand the impact of our framework's components on its performance. We also conducted extensive research analyzing our framework using various capture devices.

\textbf{Experiment I: \textit{Intra-Capture} Device}.

In this study, we assess the performance of PAD systems by training and testing them on finger photos obtained from the same capture device. 

From Table \ref{intra}, we can see that the \textit{ColFigPhotoAttnNet} outperforms all the baseline models across tested devices with BPCER of 20.24\% and 6.73\% for Google, 12.78\%, and 6.21\% for iPhone, 1.11\% and 0.21\% for OPO, and 0.11\% and 0.19\% for Nokia at APCERs of 5\% and 10\%, respectively, suggesting it is the most robust among the tested models. The MobileNet-V3 Large variant shows BPCER of 35.03\% and 14.01\% for Google and 34.51\% and 23.22\% for iPhone, indicating moderate performance. OPO and Nokia are lower at 8.05\%, 0.57\%, and 8.33\% and 1.56\%, respectively. On the other hand, MobileNet-V3 Small shows BPCER of 37.34\% and 18.84\% for Google, 36.46\% and 17.96\% for iPhone, 1.59\% and 0.53\% for OPO, and 2.12\% and 0.53\% for Nokia. Notably, the performance is better at an APCER of 10\% for iPhone, OPO, and Nokia, while an APCER of 5\% shows better results for OPO and Nokia when compared to the large variant. 

The Mobile VIT-XS and Mobile VIT-XXS models show varying BPCER, especially with iPhone captures where Mobile VIT-XS records 62.97\% and 50.14\%, and Mobile VIT-XXS records 53\% and 40.55\%. EfficientNet B7’s performance is quite variable, with BPCER as low as 1.22\% and 0.52\% for OPO, but up to 26.81\% and 17.39\% for iPhone. EfficientNet B5, meanwhile, shows a more balanced BPCER across devices, with its highest being 27.11\% and 18.01\% for iPhone and the lowest at 2.56\% and 1.03\% for OPO. The Swin Transformer model performs competitively with MobileNets, showing BPCER of 1.64\% and 0.27\% with OPO as the capture device and 1.05\% and  and 0.27\%  with OPO as the capture device and 1.05\%  and 0.52\% important discussions:

\begin{itemize} 

\item There is considerable variation in model performance based on the capture device. For instance, all models perform better on OPO and Nokia capture devices, while the error of Google and iPhone is higher overall. 

\item Google Pixel 3 and iPhone 13 Pro are known for their high-quality cameras. This could suggest that the models face more challenges in identifying attacks when captured from higher-end cameras.

\item From Fig.\ref{roc1}, we can observe that the MobileViT XS and XSS are performing poorly on all the devices, which could be attributed to the miniature size of these models compared to others.


\end{itemize}

\textbf{Experiment II: \textit{Inter-Capture} Device}.

In this study, we assess the performance of PAD systems by training and testing them on finger photos obtained from different capture devices. Table \ref{inter} presents the results of different models in \textit{\textit{inter-capture}} variation settings. The \textit{ColFigPhotoAttnNet} shows lower BPCER @ APCER = 5\% compared to other state-of-the-art models in most cases. Notably, when our model is trained on Google and tested on iPhone, it achieves a BPCER @ APCER = 10\% of 26.09\%. Similarly, 4.812\% when trained on OPO and tested on Nokia, and 1.03\% when trained on Nokia and tested on OPO. From Table \ref{inter}, we observe some trends worth discussing. For example, EfficientNet-B7 shows a BPCER @ APCER = 5\% of 7.03\%, whereas \textit{ColFigPhotoAttnNet} has a BPCER @ APCER = 5\% of 11.22\%. Although the model performs better in this scenario, it is important to note that the EfficientNet-B7 is approximately three times larger, with 66.34M parameters. EfficientNet-B5 is also competitive when trained on OPO and tested on Nokia, achieving a BPCER of 9.49\% and 3.31\% when APCER is 5\% and 10\%, respectively.


However, \textit{ColFigPhotoAttnNet} performs better at APCER = 5\%, with a BPCER of 11.22\%, compared to MobileNet-V3 Small's BPCER of 12.5\% under the same conditions. Additionally, model shows exceptional performance when trained on Nokia capture devices and tested on OPO, achieving BPCER of 3.09\% and 1.03\% at APCERs of 5\% and 10\%, respectively. The Swin Transformer model showcases variable performance across different devices. In particular, the model performs well with BPCER @ APCER = 5\% of 26.42\% and 11.64\% when trained on OPO and Nokia and tested on Nokia and OPO, respectively. A clear trend emerges that all the above models have high BPCER and do not generalize well between Google Pixel and iPhone-13 \textit{inter-capture} devices.

The following observations are made from the study: 

\begin{itemize} 
\item A trend can be observed when performing inter-device experiments, such as when trained on Nokia and tested on OPO, suggesting that the model can generalize well with older devices. However, all the models struggle when generalizing between Google Pixel 3 and iPhone-13, which indicates that as technology increases, it is difficult for models to generaliz 

\item MobileNet-V3 small also has exceptional performance when trained on OPO and tested on Nokia (3.12\% when APCER = 10\%). This indicates that smaller models can still achieve good results in older devices. Also, some models, such as EfficientNet-B7 and EfficientNet-B5, show competitive performance on older capture devices.

\item In Fig.\ref{boxplot}, in the \textit{inter-capture} scenario, the performance of the models is not consistent when dealing with capture devices that were not trained. In all the cases, we can  observe a higher median error rate.On the other hand, all the models tend to perform best when both trained and tested on the same device. 

\item Fig.\ref{boxplot} also illustrates that the \textit{ColFigPhotoAttnNet} can generalize better than other baseline models across different capture devices in \textit{inter-capture} scenarios, as indicated by its relatively lower median. Also, the smaller box plot size indicates low variance in the BPCER values. This means that the framework consistently performs well across the different inter-class conditions with minimal fluctuations.

\end{itemize}

\textbf{Ablation Study}.

\begin{figure}[]
\centering
{\includegraphics[width=7.8cm,height=6.8cm]{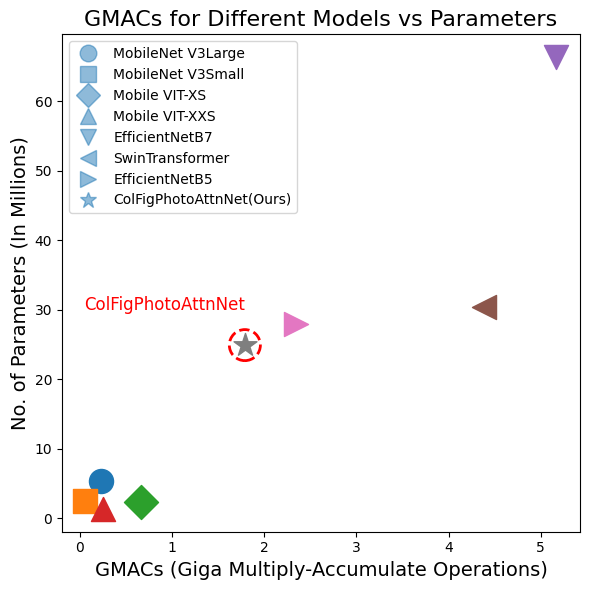}}
\caption{No. of parameters vs GMACs}
\centering
\label{GMAC}
\end{figure}

Table \ref{ablation} presents the results of an ablation study only on "Database 1" aimed at evaluating the effect of different color spaces (RGB, HSV, YCbCr), the utilization of MobileNet-V3 large as a feature extractor, the implementation of window attention mechanisms and nested residual blocks on the performance of a CNN-based model. Additional results on ablation study are presented in the supplementary paper. 

The ColFigPhotoAttnNet model achieves a BPCER of 16.77\% at APCER 5\% and 10.94\% at APCER 10\% when combining RGB color space with MobileNet as the feature extractor, alongside the bottleneck window attention and the nested residual block. Improvement is noted when the model incorporates RGB and HSV color spaces, where the BPCER drops slightly to 14.12\% at APCER 5\% and 7.868\% at APCER 10\%. The best performance is recorded at a BPCER of 9.32\% at APCER 5\% and 5.03\% at APCER 10\% when the model utilizes all three color spaces (RGB, HSV, YCbCr) while integrating MobileNet, bottleneck window attention, and the nested residual block. After the quantization block, the model’s BPCER increased to 12.78\% at APCER 5\% and 6.21\% at APCER 10\%.
In contrast, the highest BPCER at APCER 5\% of 21.79\% is observed when all color spaces are utilized without the nested residual block, indicating a significant performance decrease. Additionally, removing the bottleneck window attention from the configuration leads to a BPCER of 24.16\% at APCER 5\% and 12.92\% at APCER 10\%. The following observations are evident from ablation study: 



\begin{itemize} 
\item The performance of the model improved when different color spaces were used, suggesting that these color spaces capture and represent various aspects of color information as distinct features. 

\item The attention mechanism in convolutional layers adds weight to specific image regions, enhancing recognition, while nested residual blocks with skip connections reduce vanishing gradients, allowing deeper feature learning.

\item During the dynamic quantization, the model’s error rate increased from a BPCER of 9.32\% at APCER 5\% and 5.03\% at APCER 10\% to 12.78\% at APCER 5\% and 6.21\% at APCER 10\% but significantly reduced the model’s parameters from 32.2 million to 24.89 million (by 7.3 million). This demonstrates a calculated trade-off between model compactness and robustness.

\item The \textit{ColFigPhotoAttnNet} model demonstrates an efficient balance between the number of parameters and computational complexity. Fig \ref{GMAC} shows a balance which suggests that it can achieve competitive performance while maintaining efficiency, making it an attractive option for PAD applications where computational resources are limited.

\item \textit{ColFigPhotoAttnNet} has a Giga Multiply-Accumulate Operations (GMAC) value of 1.79 and Parameters are 24.89M which is moderate between models like EfficientNet-variants and Swin Transformer and smaller models like Mobile variants. More information is provided in the supplementary.
\end{itemize}

\section{Conclusion}

This is the first study to examine how different capture devices affect finger photo PAD systems. This paper introduces a hybrid \textit{ColFigPhotoAttnNet} framework, incorporating window self-attention layers focused on distinct color spaces and nested residual connections. \textit{ColFigPhotoAttnNet} shows better performance in \textit{intra-capture} and \textit{inter-capture} surpassing state-of-the-art models in generalization. We can see that the model achieved an BPCER @ APCER= 5\% of 0.11\% on Nokia, 1.11\% on OPO, 12.78\% on iPhone, and 20.24\% on Google databases. We also observed a trend in inter-capture scenarios, indicating that newer technology based capture devices have a significant impact on PAD which is more difficult to generalize. 

In the ablation results, we can observe how various components affect the performance of PAD. After applying dynamic quantization, the model's BPCER @ APCER=5\% increased from 9.32\% to 12.78\% but significantly reduced the model's parameters by 7.3 million. Additionally, we notice a trade-off between model performance and parameter size, and quantization effectively addresses this challenge.


{\small
\bibliographystyle{ieee_fullname}
\bibliography{PaperForReview}
}

\end{document}


\title{Supplementary- ColFigPhotoAttnNet: Reliable Finger Photo Presentation Attack Detection Leveraging Window-Attention on Color Spaces}

\author{
Anudeep Vurity\textsuperscript{1}, Emanuela Marasco\textsuperscript{1}, Raghavendra Ramachandra\textsuperscript{2}, Jongwoo Park\textsuperscript{3} \\
\\
\textsuperscript{1}Center for Secure and Information Systems, George Mason University, U.S.A.\\
\textsuperscript{2}Norwegian University of Science and Technology (NTNU), Gjøvik, Norway\\
\textsuperscript{3}Stony Brook University, Stony Brook, NY, U.S.A\\
{\tt\small \{avurity, emarasco\}@gmu.edu, raghavendra.ramachandra@ntnu.no, jongwopark@cs.stonybrook.edu}
}

\maketitle

\begin{figure*}[h]
    \small
    \centering
    \begin{subfigure}[b]{0.45\textwidth}
        \centering
        \includegraphics[width=\textwidth]{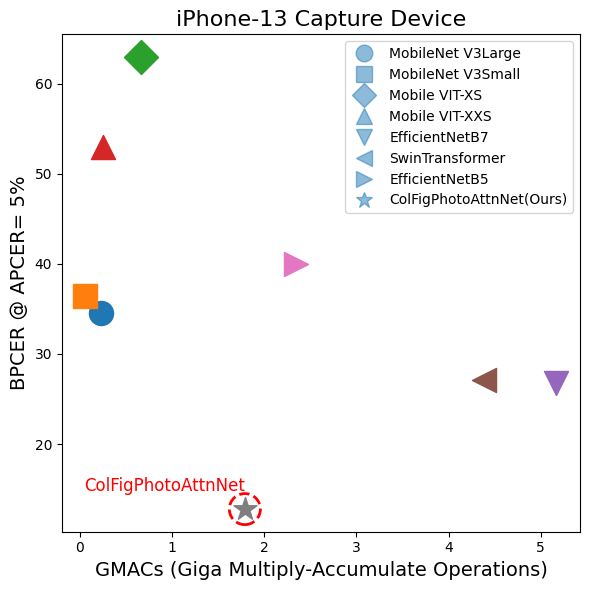}
        \caption{GMACs vs Model Performance: iPhone13 capture device}
        \label{fig:gmac1}
    \end{subfigure}
    \hfill
    \begin{subfigure}[b]{0.45\textwidth}
        \small
        \centering
        \includegraphics[width=\textwidth]{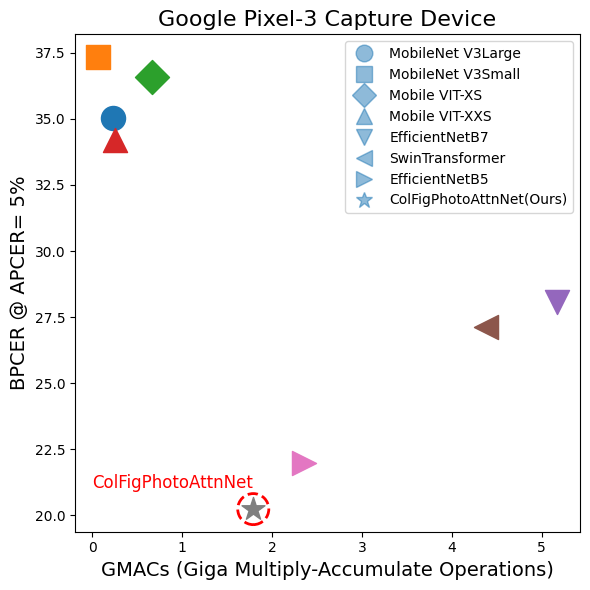}
        \caption{GMACs vs Model Performance: Google Pixel-3 capture device}
        \label{fig:gmac2}
    \end{subfigure}
    \hfill
    \begin{subfigure}[b]{0.45\textwidth}
        \small
        \centering
        \includegraphics[width=\textwidth]{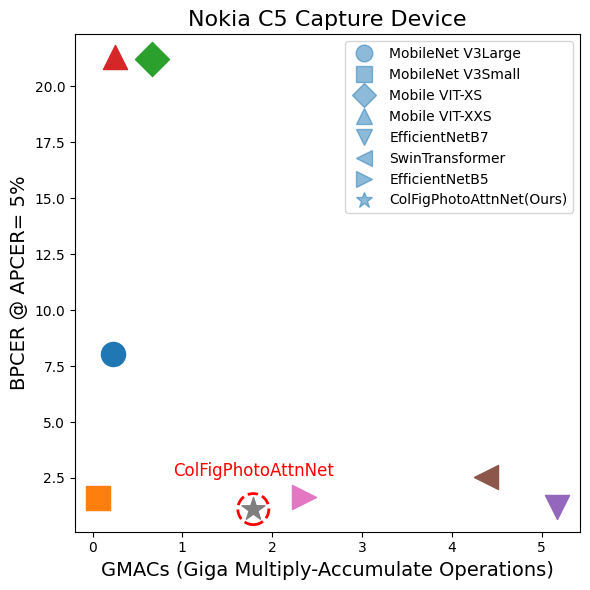}
        \caption{GMACs vs Model Performance: Nokia C5 capture device}
        \label{fig:gmac3}
    \end{subfigure}
    \hfill
    \begin{subfigure}[b]{0.45\textwidth}
        \small
        \centering
        \includegraphics[width=\textwidth]{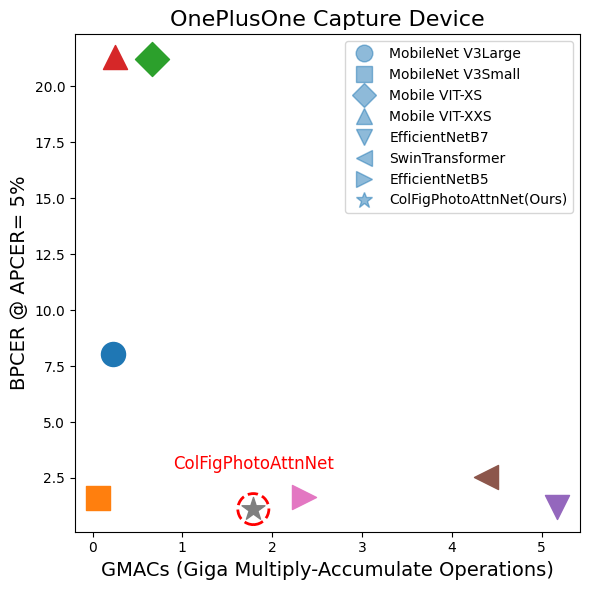}
        \caption{GMACs vs Model Performance: OnePlusOne capture device}
        \label{fig:gmac4}
    \end{subfigure}
    \caption{The figure shows complexity vs Performance of the various models on all the different capture devices. ColFigPhotoAttnNet is highlighted with a red circle. a,b,c,d clearly shows that \textit{ColFigPhotoAttnNet} shows performance vs complexity trade-off. }
    \label{fig:all_gmac}
\end{figure*}

\section{Giga Multiply-Accumulate Operations}

Giga Multiply-Accumulate Operations (GMAC), is a metric for computational complexity. It is used to measure the number of multiplication and addition operations a model makes during the phase of inference. This is an important metric to have, as it shows how computationally burdensome a model would be. GMACs provides a standard way to compare models based on their efficiency, especially in deep learning, where computations are intensive.

\subsection{Explanation of GMACs for Different Models}
The GMAC values for various models used in this paper are presented as follows:

GMAC values represents the computational complexity of each model. Table.\ref{baseline_models} shows all the Complexity of models and number of parameters used in this research. For instance, models like EfficientNet-B7 and SwinTransformer have high GMAC values: 5.17 and 4.38, respectively. Therefore, they have higher requirements for computational power, which, in turn, may give better performance but increase resource consumption. By contrast, models such as MobileNet-V3 Small (0.061 GMACs) and Mobile VIT-XXS (0.254 GMACs) have low GMACs—mostly used in scenarios where computational resources are limited. 

Fig.\ref{fig:all_gmac} shows the performance of all the applied models and the model complexity on each capture devices. \textit{ColFigPhotoAttnNet} exhibits a balance between computational cost and performance. The figures depict the relationship between GMACs and BPCER (BonaFide Presentation Classification Error Rate) at a 5\% APCER (Attack Presentation Classification Error Rate) for various models across iPhone, Google, Nokia and OPO capture devices.  

The GMAC value for the \textit{ColFigPhotoAttnNet} model is 1.79. Although \textit{ColFigPhotoAttnNet} is more computationally demanding than some models, such as MobileNet-V3 Small, it was far less demanding compared with models like EfficientNet-B7 and Swin Transformer. This suggests that ColFigPhotoAttnNet is balanced between computational complexity and performance, which is suitable for applications that require fewer computational resources.

\begin{table}[h]
\begin{center}
\renewcommand{\arraystretch}{1.2}
\setlength{\tabcolsep}{10pt}
\footnotesize

\begin{tabular}{|l|c|c|}
\hline
\textbf{Baseline Models} & \textbf{GMACs} & \textbf{Parameters (M)} \\ \hline
MobileNet V3Large & 0.231 & 5.4 \\ \hline
MobileNet V3Small & 0.061 & 2.5 \\ \hline
Mobile VIT-XS & 0.665 & 2.3 \\ \hline
Mobile VIT-XXS & 0.254 & 1.3 \\ \hline
EfficientNetB7 & 5.17 & 66.34 \\ \hline
SwinTransformer & 4.38 & 27.89 \\ \hline
EfficientNetB5 & 2.35 & 30.38 \\ \hline
\textbf{ColFigPhotoAttnNet (Ours)} & 1.79 & 24.89 \\ \hline
\end{tabular}

\end{center}
\caption{Baseline models, GMACs, and Parameters}
\label{baseline_models}
\end{table}

\section{Additional Ablation Results}

In this supplementary, we show the results of ablation studies on different datasets used. The Table.\ref{ablation1} presents an ablation study using Google Pixel 3 capture device to evaluate the effectiveness of different model configurations in terms of BPCER at APCER thresholds: 5\% and 10\%. When only the RGB color space is used, the BPCER values are relatively high, with 22.91\% at APCER 5\% and 11.55\% at APCER 10\%. Adding the HSV color space alongside RGB reduces the BPCER to 15.64\% and 7.45\%, respectively. Including the YCbCr color space with RGB results in BPCER values of 17.58\% and 8.64\%. The inclusion of a bottleneck attention mechanism improves the model's performance. When using all three color spaces with bottleneck attention but without residual blocks yields a BPCER of 18.24\% at APCER 5\% and 5.91\% at APCER 10\%. 

\begin{table}[]
\renewcommand{\arraystretch}{1.2}
\setlength{\tabcolsep}{2.5pt}
\footnotesize

\begin{tabular}{|ccc|c|c|c|c|c|}
\hline
\multicolumn{3}{|c|}{\textbf{Color Spaces}} & \textbf{BottleNeck} & \multirow{2}{*}{\textbf{\begin{tabular}[c]{@{}c@{}}Residual \\ Block\end{tabular}}} & \textbf{DQ} &\textbf{BPCER} & \textbf{BPCER} \\ \cline{1-4} \cline{7-8} 
\multicolumn{1}{|c|}{\textbf{RGB}} & \multicolumn{1}{c|}{\textbf{HSV}} & \textbf{YCbCr} & \textbf{Attention} &  &  & \textbf{5\%} & \textbf{10\%} \\ \hline
\multicolumn{1}{|c|}{\checkmark} & \multicolumn{1}{c|}{x} & x & \checkmark & \checkmark & x & 22.91 & 11.55 \\ \hline
\multicolumn{1}{|c|}{\checkmark} & \multicolumn{1}{c|}{\checkmark} & x & \checkmark & \checkmark & x & 15.64 & 7.45 \\ \hline
\multicolumn{1}{|c|}{\checkmark} & \multicolumn{1}{c|}{x} & \checkmark & \checkmark & \checkmark & x & 17.58 & 8.64 \\ \hline
\multicolumn{1}{|c|}{\checkmark} & \multicolumn{1}{c|}{\checkmark} & \checkmark & \checkmark & x & x & 28.95 & 13.38 \\ \hline
\multicolumn{1}{|c|}{\checkmark} & \multicolumn{1}{c|}{\checkmark} & \checkmark & x & \checkmark & x & 31.48 & 20.64 \\ \hline
\multicolumn{1}{|c|}{\checkmark} & \multicolumn{1}{c|}{\checkmark} & \checkmark & \checkmark & \checkmark & x & 18.24 & 5.91 \\ \hline
\multicolumn{1}{|c|}{\checkmark} & \multicolumn{1}{c|}{\checkmark} & \checkmark & \checkmark & \checkmark & \checkmark & 20.24 & 6.73 \\ \hline
\end{tabular}
\caption{Ablation study using Database 2, This table shows BPCER @APCER 5\% and APCER 10\%.} 
\label{ablation1}
\end{table}

\begin{table}[]
\begin{center}
\renewcommand{\arraystretch}{1.2}
\setlength{\tabcolsep}{2.5pt}
\footnotesize

\begin{tabular}{|ccc|c|c|c|c|c|}
\hline
\multicolumn{3}{|c|}{\textbf{Color Spaces}} & \textbf{BottleNeck} & \multirow{2}{*}{\textbf{\begin{tabular}[c]{@{}c@{}}Residual \\ Block\end{tabular}}} & \textbf{DQ} &\textbf{BPCER} & \textbf{BPCER} \\ \cline{1-4} \cline{7-8} 
\multicolumn{1}{|c|}{\textbf{RGB}} & \multicolumn{1}{c|}{\textbf{HSV}} & \textbf{YCbCr} & \textbf{Attention} &  &  & \textbf{5\%} & \textbf{10\%} \\ \hline
\multicolumn{1}{|c|}{\checkmark} & \multicolumn{1}{c|}{x} & x & \checkmark & \checkmark & x & 1.24 & 0.25 \\ \hline
\multicolumn{1}{|c|}{\checkmark} & \multicolumn{1}{c|}{\checkmark} & x & \checkmark & \checkmark & x & 1.05 & 0.14 \\ \hline
\multicolumn{1}{|c|}{\checkmark} & \multicolumn{1}{c|}{x} & \checkmark & \checkmark & \checkmark & x & 0.64 & 0.51 \\ \hline
\multicolumn{1}{|c|}{\checkmark} & \multicolumn{1}{c|}{\checkmark} & \checkmark & \checkmark & x & x & 3.74 & 1.41 \\ \hline
\multicolumn{1}{|c|}{\checkmark} & \multicolumn{1}{c|}{\checkmark} & \checkmark & x & \checkmark & x & 1.21 & 0.21 \\ \hline
\multicolumn{1}{|c|}{\checkmark} & \multicolumn{1}{c|}{\checkmark} & \checkmark & \checkmark & \checkmark & x & 0.00 & 0.00 \\ \hline
\multicolumn{1}{|c|}{\checkmark} & \multicolumn{1}{c|}{\checkmark} & \checkmark & \checkmark & \checkmark & \checkmark & 0.11 & 0.19 \\ \hline
\end{tabular}
\end{center}
\caption{Ablation study using Nokia Capture Device, This table shows BPCER @APCER 5\% and APCER 10\%.} 
\label{ablation2}
\end{table}

\begin{table}[]
\begin{center}
\renewcommand{\arraystretch}{1.2}
\setlength{\tabcolsep}{2.5pt}
\footnotesize

\begin{tabular}{|ccc|c|c|c|c|c|}
\hline
\multicolumn{3}{|c|}{\textbf{Color Spaces}} & \textbf{BottleNeck} & \multirow{2}{*}{\textbf{\begin{tabular}[c]{@{}c@{}}Residual \\ Block\end{tabular}}} & \textbf{DQ} &\textbf{BPCER} & \textbf{BPCER} \\ \cline{1-4} \cline{7-8} 
\multicolumn{1}{|c|}{\textbf{RGB}} & \multicolumn{1}{c|}{\textbf{HSV}} & \textbf{YCbCr} & \textbf{Attention} &  &  & \textbf{5\%} & \textbf{10\%} \\ \hline
\multicolumn{1}{|c|}{\checkmark} & \multicolumn{1}{c|}{x} & x & \checkmark & \checkmark & x & 6.91 & 2.85 \\ \hline
\multicolumn{1}{|c|}{\checkmark} & \multicolumn{1}{c|}{\checkmark} & x & \checkmark & \checkmark & x & 3.53 & 1.45 \\ \hline
\multicolumn{1}{|c|}{\checkmark} & \multicolumn{1}{c|}{x} & \checkmark & \checkmark & \checkmark & x & 4.85 & 1.24 \\ \hline
\multicolumn{1}{|c|}{\checkmark} & \multicolumn{1}{c|}{\checkmark} & \checkmark & \checkmark & x & x & 1.08 & 0.54 \\ \hline
\multicolumn{1}{|c|}{\checkmark} & \multicolumn{1}{c|}{\checkmark} & \checkmark & x & \checkmark & x & 1.81 & 1.62\\ \hline
\multicolumn{1}{|c|}{\checkmark} & \multicolumn{1}{c|}{\checkmark} & \checkmark & \checkmark & \checkmark & x & 0.64 & 0.00 \\ \hline
\multicolumn{1}{|c|}{\checkmark} & \multicolumn{1}{c|}{\checkmark} & \checkmark & \checkmark & \checkmark & \checkmark & 1.11 & 0.21 \\ \hline
\end{tabular}
\end{center}
\caption{Ablation study using OPO Capture Device, This table shows BPCER @APCER 5\% and APCER 10\%.} 
\label{ablation3}
\end{table}

\begin{figure}[]
{\includegraphics[width=9cm,height=7.2cm]{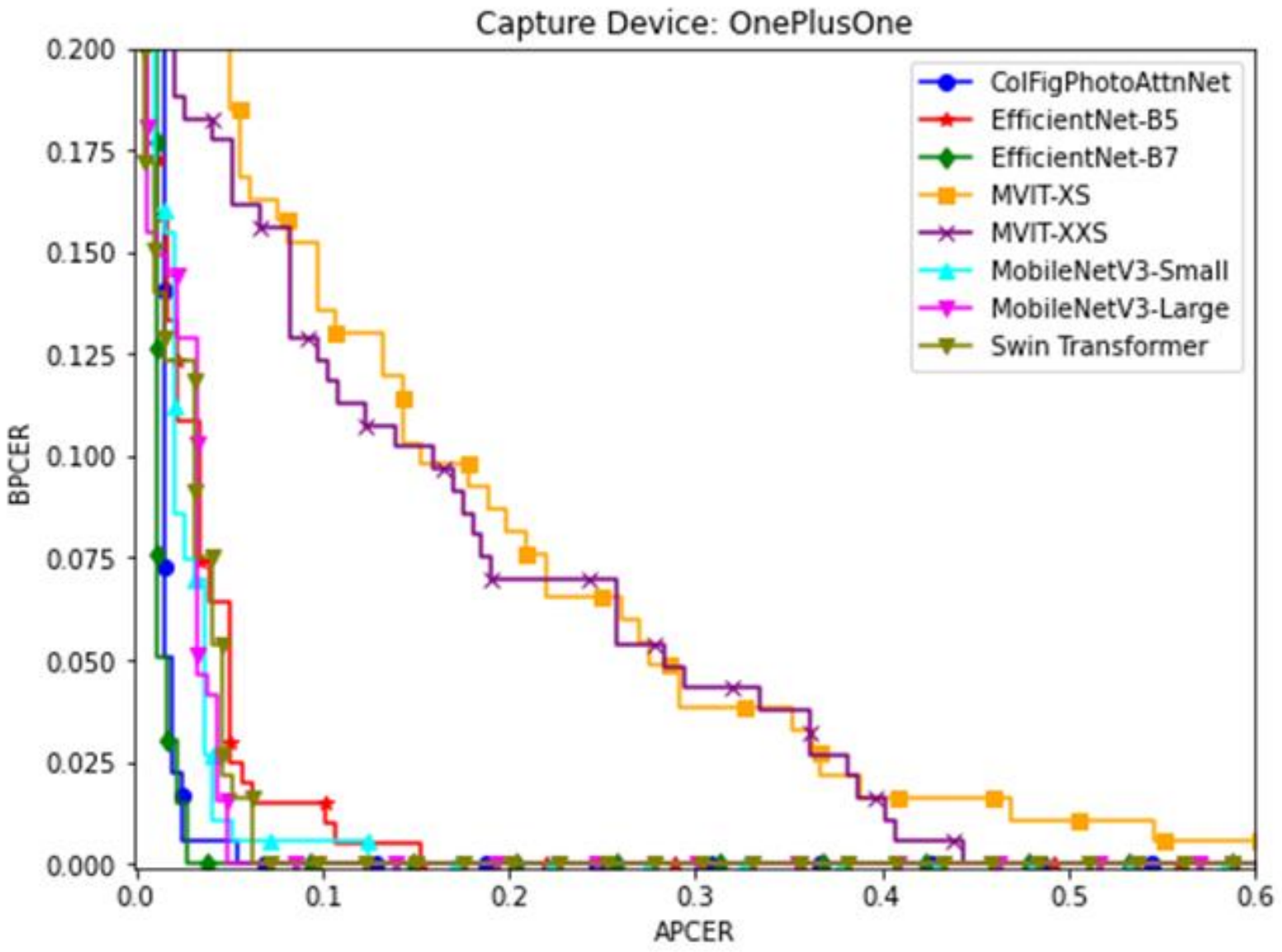}}
\caption{OPO capture device}
\centering
\label{opo}
\end{figure}

\begin{figure}[]
{\includegraphics[width=9cm,height=7.2cm]{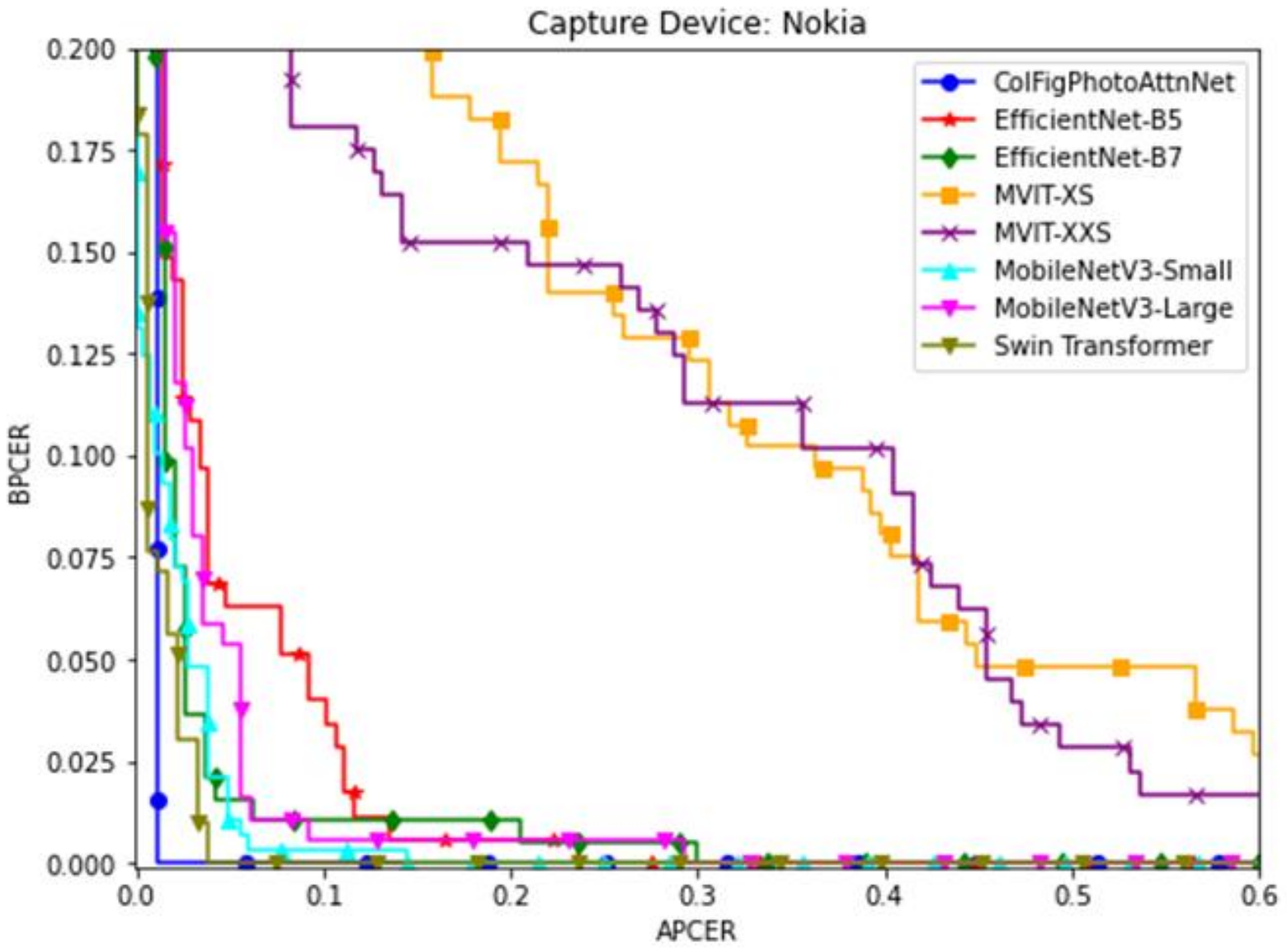}}
\caption{Nokia capture device}
\centering
\label{nokia}
\end{figure}

Similarly, in Table.\ref{ablation2} presents an ablation study using Nokia capture device and Table \ref{ablation3} presents ablation study using OPO capture device. For both devices, the use of multiple color spaces generally improves model performance. The inclusion of bottleneck attention and residual blocks further enhances the model's accuracy. When tested on Nokia device, the configuration with all three color spaces, bottleneck attention, and residual blocks achieves a BPCER of 0.00\% at both APCER 5\% and 10\%. Similarly, on the OPO device, the same configuration results in BPCER values of 0.64\% and 0.00\%, respectively.  Figures \ref{opo} and \ref{nokia} illustrate the performance of various models on Database 1 when using OPO and Nokia capture devices, respectively. The results clearly demonstrate that \textit{ColFigPhotoAttnNet} model  outperforms the baseline models in both scenarios.

Applying dynamic quantization (DQ) is found to reduce the model's complexity, although it slightly decreases performance. For instance, the configuration with DQ on the Nokia device shows a slight increase in BPCER compared to without DQ (0.11\% vs. 0.00\% at APCER 5\%). On the OPO device, a similar trend is observed, with BPCER values of 1.11\% at APCER 5\% and 0.21\% at APCER 10\% when DQ is applied.

R2-Q2: Secondly, I would suggest that you present your results in a more balanced manner. While it is impressive that ColFigPhotoAttnNet outperforms other models on some capture devices, the lack of convincing evidence of its generalization ability across different datasets and scenarios is concerning. I would recommend including additional experiments or using existing datasets to demonstrate the model’s robustness and ability to generalize.

Yes, Efficient-b7, Efficient-b5 has shown better perofrmance on OPO dataset on intra class study. Therefore, as reviewer suggested We added additional intra class results Leave One Out method.   

The results and observations are given below:

\begin{itemize}
    \item  The inclusion of HSV and YCbCr alongside RGB consistently shows better results compared to using RGB alone. When using multiple color spaces improves model performance. Configurations that include RGB, HSV, and YCbCr tend to yield lower BPCER values at both APCER thresholds (5\% and 10\%).
    \item The inclusion of bottleneck attention mechanisms significantly enhances model performance across all devices. Residual blocks further improve performance when used in conjunction with bottleneck attention. Models with both bottleneck attention and residual blocks achieve the lowest BPCER values in most cases. For instance, the configuration with all three color spaces, bottleneck attention, and residual blocks yields the best results on both the Nokia and OPO devices. 
    \item Dynamic Quantization (DQ) decreases the model complexity but marginally increases the BPCER\%. Although there is a minor trade-off in BPCER\%, this reduction in model complexity can be very useful for real-world applications with limited computational power.
\end{itemize}

\section{Additional inter-capture results} 
This t-SNE plots (Fig.\ref{tsne} \ref{tsne1}) illustrates the inter- and intra-device generalization capabilities of most competing models, focusing on both old and new datasets. In this analysis, we use one new capture device (Google) and one older dataset (OnePlusOne) to evaluate how well ColFigAttnNet generalizes across devices. Specifically, we assess the model's performance when training and testing occur on the same device (intra-scenario) and when testing is performed on a different device (inter-scenario). EfficientNet-b5 and EfficientNet-b7 show competitive performance alongside ColFigAttnNet, prompting us to further investigate how these models generalize in similar scenarios. The t-SNE plot highlights the challenge of separating the live (blue) and spoof (orange) classes, with models showing varying levels of separation across different devices and testing conditions.

\subsection{t-SNE analysis: Google Train} 

\begin{figure}[]
\small
{\includegraphics[width=8.4cm,height=7.2cm]{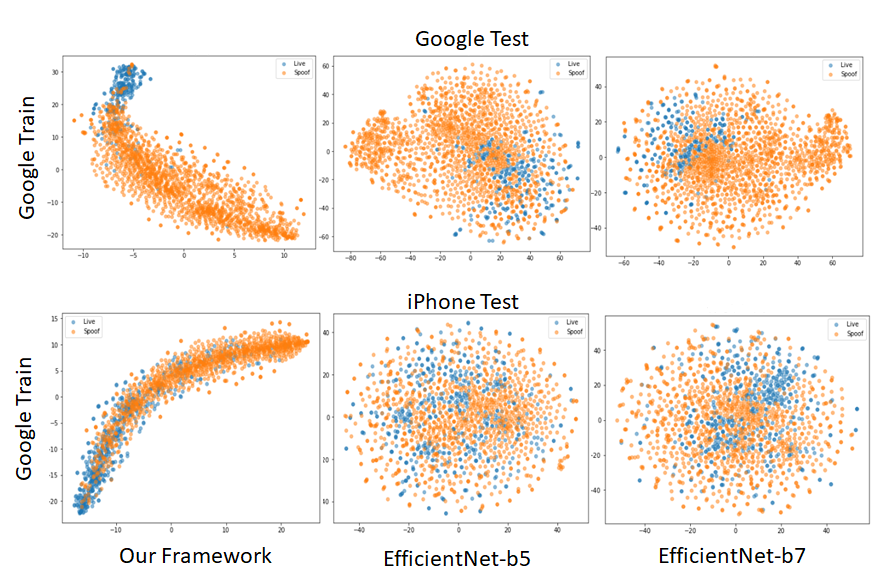}}
\caption{t-SNE plot illustrating the generalization performance in both inter- and intra-device scenarios for models trained on the Google dataset}
\centering
\label{tsne}
\end{figure}

\begin{itemize}
    \item The Fig.\ref{tsne} shows clear separation between live and spoof samples in the Google Train and Google Test scenario.This indicates that the model performs well in the intra-device scenario, learning features that generalize well when both training and testing are on the same device (Google).

    \item The separation is good even in the inter-device scenario, where the model was trained on Google but tested on iPhone. While the points are more dispersed, the separation is still visible, indicating that the framework has better generalization.

    \item EfficientNet b5 and b7 shows a heavy overlap between live and spoof classes in the iPhone Test set, showing that EfficientNet-b5 has difficulty in generalizing from Google Train to the iPhone Test. This inter-device scenario proves challenging for the model.

    \item There is significant overlap between live and spoof samples in the inter scenerio for both efficientNet-b5 and b7, indicating that both struggles to generalize in inter scenario. 

\begin{figure}[]
\small
{\includegraphics[width=8.4cm,height=6cm]{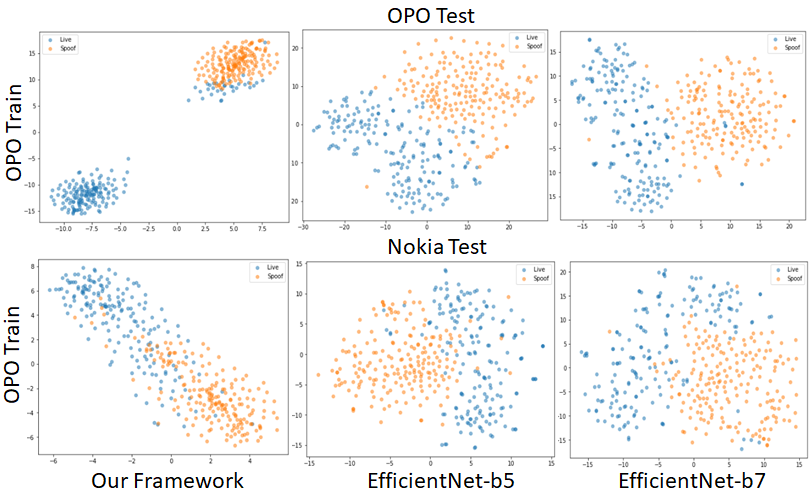}}
\caption{t-SNE plot illustrating the generalization performance in both inter- and intra-device scenarios for models trained on the OPO dataset.}
\centering
\label{tsne1}
\end{figure}
   
\end{itemize}

\subsection{t-SNE analysis: OPO Train}

\begin{itemize}
    \item Fig.\ref{tsne1} shows that live and spoof classes are clearly well-defined in two different clusters. This indicates that the model performs exceptionally well when both the training and testing data come from the same OPO capture device.
    
    \item EfficientNet-b5 and EfficientNet-b5 shows weaker separations between live and spoof when compared to ColFigPhotoAttnNet in intra-scenario.

    \item In inter class scenario where the model was trained on OPO but tested on Nokia. We can see clusters are more dispersed compared to the intra-device scenario. Although the model still maintains a good separation between live and spoof classes, demonstrating its strong generalization ability across devices.
    \item In inter-class scenerio, although all the models have moderate saperation, EfficeintNet-B5 and B7 separation has noticeable overlap between the live and spoof classes compared to our framework. Both EfficientNet b5 and b7 struggles more in this inter-device scenario, indicating that there is some difficulty in generalization.
\end{itemize}

From the above analysis, we can see that our Framework is more robust in both intra-device and inter-device scenarios compared to other top models, showing better separation and generalization across devices. Additional investigation will be conducted in future work to provide a more comprehensive understanding of the model's effectiveness in diverse environments.
